\documentclass[lettersize,journal]{IEEEtran}
\usepackage{amsmath,amsfonts}
\usepackage{algorithmic}
\usepackage{algorithm}
\usepackage{mathtools}
\usepackage[switch]{lineno} 

\usepackage{array}
\usepackage[caption=false,font=normalsize,labelfont=sf,textfont=sf]{subfig}
\usepackage{textcomp}
\usepackage{stfloats}
\usepackage{booktabs}   
\usepackage{graphicx}   
\usepackage{multirow}  
\usepackage{url}
\usepackage[table]{xcolor} 
\usepackage[table, svgnames]{xcolor}  
\definecolor{lightblue}{RGB}{230,240,255} 
\definecolor{RevisionBlue}{RGB}{0,0,0}
\newcommand{\red}[1]{\textcolor{red}{#1}}
\newcommand{\blue}[1]{\textcolor{blue}{#1}}

\usepackage{verbatim}
\usepackage{graphicx}
\usepackage{cite}
\usepackage{xcolor}
\usepackage{bm}
\usepackage{titletoc}
\usepackage[colorlinks=true, citecolor=blue]{hyperref}

\begin{document}

\title{Bridging Information Asymmetry: A Hierarchical Framework for Deterministic Blind Face Restoration}

\author{
Zhengjian Yao, Jiakui Hu, Kaiwen Li, Hangzhou He, Xinliang Zhang, Shuang Zeng, Lei Zhu, Yanye Lu
\thanks{Zhengjian Yao, Jiakui Hu, Kaiwen Li, Hangzhou He, Xinliang Zhang, Shuang Zeng, Lei Zhu and Yanye Lu are with the Biomedical Engineering Department, College of Future Technology, Peking University, Beijing, China. They are also with the Institute of Medical Technology, Peking University Health Science Center, and the National Biomedical Imaging Center, Peking University, Beijing, China.}
\thanks{Corresponding Authors: Yanye Lu (yanye.lu@pku.edu.cn).}
}



\maketitle


\begin{abstract}

Blind face restoration remains a persistent challenge due to the inherent ill-posedness of reconstructing holistic structures from severely constrained observations. Current generative paradigms, while capable of synthesizing realistic facial details, remain limited by the under-constrained nature of blind restoration, where severely degraded inputs can be mapped to plausible yet identity-inconsistent outputs. To address this issue, we present \textbf{Pref-Restore}, a hierarchical framework for deterministic BFR. Our design is organized around three complementary principles: (1) Semantic Information Augmentation, where an auto-regressive semantic branch converts image and text cues into structured tokens that provide a stable high-level anchor; (2) Texture-level Fidelity Alignment, where the diffusion generator is trained under this anchor to recover identity-relevant details; and (3) Fidelity-constrained Preference Optimization, where a face-aware reward refines the diffusion trajectory while controlling the quality--fidelity trade-off. Extensive experiments on synthetic and real-world benchmarks show that Pref-Restore achieves state-of-the-art performance, with stronger identity-sensitive fidelity and lower restoration uncertainty across repeated sampling. Systematic ablations further attribute these gains to the proposed hierarchical design, showing the necessity of staged training, the robustness of the text pathway under deployment-faithful conditions, and the benefit of fidelity-constrained preference optimization.
The code is available at \url{https://github.com/zjYao36/Pref_Restore/}.

\end{abstract}


\section{Introduction}


\IEEEPARstart{B}{lind} Face Restoration (BFR) aims to recover high-quality (HQ) facial images from low-quality (LQ) inputs corrupted by unknown degradations~\cite{wang2021gfpgan,yang2021gpen,gu2022vqfr,zhou2022codeformer,yue2024difface,wang2023dr2,tsai2023DAEFR}. Unlike general restoration tasks~\cite{you1999blind,kundur2002blind,yue2024deep,lin2023diffbir,hu2025dcpt, jiang2025survey}, {\color{RevisionBlue}BFR must recover identity-faithful facial structures from severely constrained observations. This is particularly challenging in natural scenes, where facial components often occupy limited spatial resolutions and critical high-frequency cues are heavily degraded. As a result, BFR remains a highly ill-posed inverse problem: the model should not merely synthesize a plausible face, but recover the target identity under substantial uncertainty.
}

\begin{figure}[!t]
\centering
\includegraphics[width=0.9\linewidth]{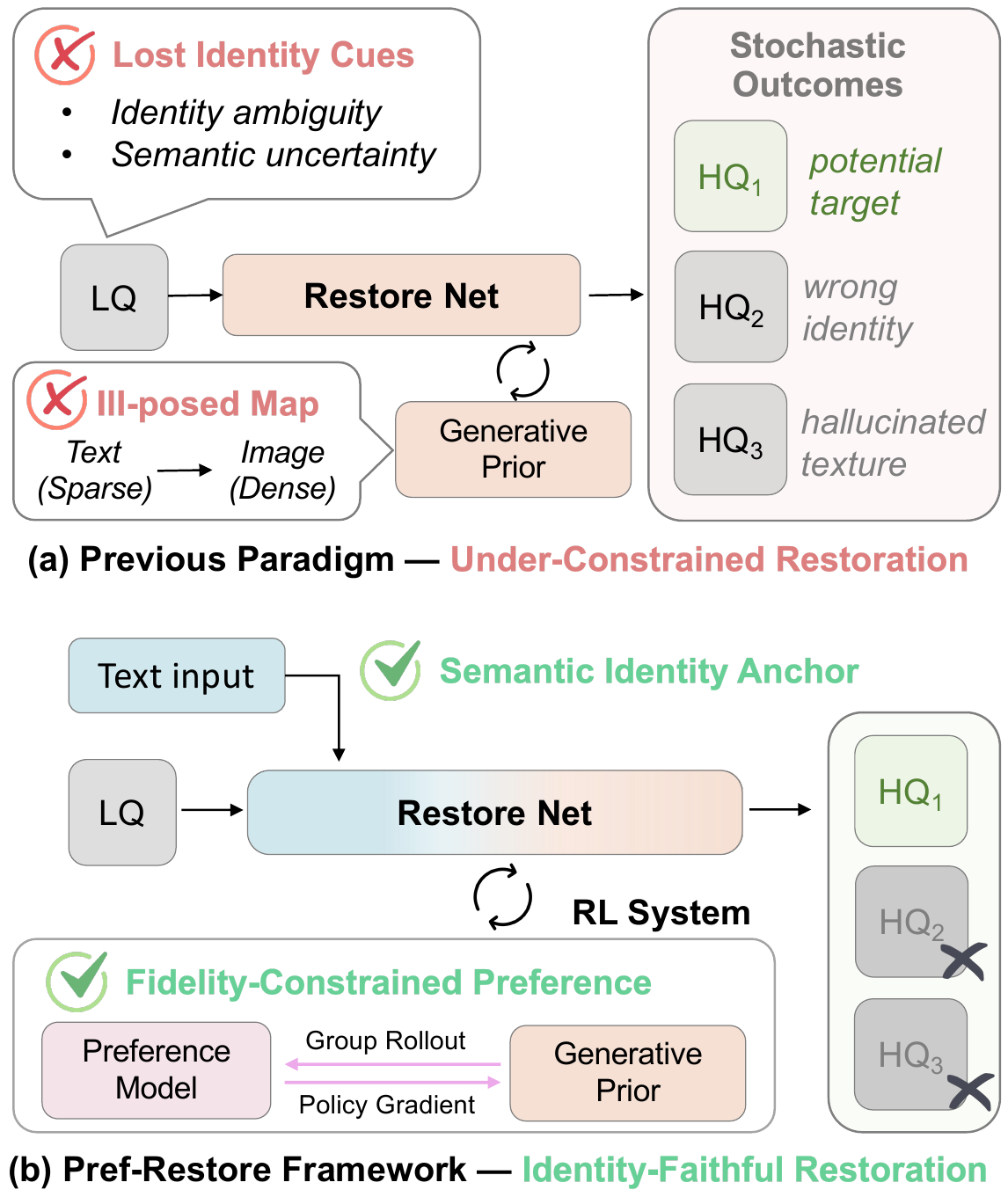}
\vspace{-1mm}
\caption{\textbf{Conceptual illustration of (a) the conventional generative prior paradigm versus (b) our proposed Pref-Restore framework.} Existing methods suffer from information asymmetry, where \textit{ill-posed priors and sparse inputs} lead to stochastic outcomes such as hallucinations or identity loss. Our Pref-Restore resolves this by \textit{augmenting input density} through AR-based semantic modeling and \textit{pruning the output distribution} via on-policy reinforcement learning, effectively pruning the uncertain solution space to achieve deterministic, preference-consistent restoration.}
\vspace{-3mm}
\label{intro}
\end{figure}

{\color{RevisionBlue}
This under-constrained setting exposes different limitations in existing restoration paradigms.}
Deterministic approaches, typically optimized via pixel-wise objectives, frequently suffer from the ``regression-to-the-mean'' effect, yielding over-smoothed results~\cite{ledig2017SRGAN,wang2018esrgan,ji2020realsr}. 
Conversely, generative priors attempt to address this under-constrained problem by restricting outputs to a learned high-quality distribution.
However, we argue that this merely shifts the ill-posedness (Fig. \ref{intro} (a)):
The priors themselves are trained on an ill-posed mapping from sparse semantic signals (such as text or class labels) to dense pixel spaces\cite{kingma2013vae,goodfellow2014gan,ho2020ddpm,song2019score-matching}.
Consequently, the restoration remains prone to stochastic uncertainty and, more critically, to hallucinations that are semantically plausible yet structurally erroneous. 
Such stochasticity limits the reliability of BFR in identity-sensitive applications, necessitating a more constrained, deterministic restoration framework.

{\color{RevisionBlue}
To address this limitation, we propose Pref-Restore, a hierarchical framework for deterministic blind face restoration (Fig.~\ref{intro} (b)). Rather than relying on a stronger generative prior alone, Pref-Restore constrains the restoration process from two complementary directions: it first stabilizes identity-relevant semantic cues before texture synthesis, and then refines the diffusion trajectory under an explicit fidelity constraint.

\textbf{Semantic-Guided Identity Stabilization.} We use an Auto-Regressive (AR) semantic branch to organize multimodal cues into identity-oriented semantic tokens via a Next-Token Prediction (NTP) paradigm~\cite{pan2025metaquery, chen2025blip3o, han2025tar, hu2025mvar, hu2025omni-view}, and align these tokens with the continuous diffusion latent space through Multi-modal Knowledge Alignment~\cite{pan2025metaquery,chen2025blip3o,lin2025uniworld}. This forms a stable high-level semantic anchor that consistently guides texture restoration toward the target identity.

\textbf{Fidelity-Constrained Preference Optimization.} We integrate On-Policy Reinforcement Learning (RL) into the diffusion trajectory~\cite{shao2024grpo,liu2025flowgrpo,zheng2025diffusionnft} and optimize it with a BFR-specific reward that couples perceptual preference with facial fidelity. Under identity and structure constraints, the Preference-Aware Fine-tuning stage improves visual quality while reducing drift toward visually plausible but identity-inconsistent faces. It therefore makes the quality--fidelity balance explicitly controllable for deterministic restoration.
}

The main contributions of this work are summarized as follows:
\begin{itemize}
    \item {\color{RevisionBlue}\textbf{Conceptual Framework:} We propose Pref-Restore as a hierarchical framework that constrains BFR from two directions: input-side semantic anchoring stabilizes identity cues, while output-side preference alignment guides restoration toward fidelity-preserving solutions. This reduces identity drift and restoration uncertainty.}

    \item {\color{RevisionBlue}\textbf{Methodological Innovation:} We introduce staged semantic--texture alignment and a face-aware restoration reward for preference optimization. Systematic ablations isolate the effects of staged training, realistic text guidance, and fidelity-constrained preference optimization.}

    \item {\color{RevisionBlue}\textbf{SOTA \& Controllability:} Pref-Restore achieves state-of-the-art performance on synthetic and real-world benchmarks, with two variants offering a controllable quality--fidelity point that improves perceptual quality while preserving identity-sensitive restoration fidelity.}
\end{itemize}

\section{Related Works}

\subsection{Blind Face Restoration with Generative Priors}

Blind face restoration is an inherently ill-posed problem that necessitates strong auxiliary priors to regularize the solution space, given the unavailability of degradation kernels and identity information. 
Early attempts primarily exploit geometric priors, such as facial landmarks~\cite{chen2018fsrnet}, parsing maps~\cite{chen2021progressive}, or 3D shapes~\cite{hu2020face}, and reference priors~\cite{dogan2019exemplar,li2020blind} from guided identity images. 
However, these methods often falter in real-world scenarios, as estimating accurate geometry from severely degraded inputs is challenging, and high-quality reference images are rarely accessible.

With the advent of deep generative models, recent research has shifted towards leveraging pre-trained generative priors for their superior detail synthesis capabilities. 
Pioneering works utilizing \textit{StyleGAN}~\cite{karras2019stylegan}, such as \textit{GPEN}~\cite{yang2021gpen} and \textit{GFPGAN}~\cite{wang2021gfpgan}, incorporate structural cues from low-quality inputs to guide the generation process via GAN inversion~\cite{gu2020image,menon2020pulse} or spatial feature modulation. 
To further enhance fidelity, codebook-based priors (e.g., \textit{VQFR}~\cite{gu2022vqfr} and \textit{CodeFormer}~\cite{zhou2022codeformer}) employ vector-quantized dictionaries to retrieve high-quality texture codes. 
More recently, diffusion models have also been adapted as refinement modules (e.g., \textit{DR2}~\cite{wang2023dr2}, \textit{DiffBIR}~\cite{lin2023diffbir}) to hallucinate high-frequency details. 
Despite their impressive perceptual quality, these generative approaches fundamentally rely on the accurate mapping from degraded features to the high-quality manifold. 
This mapping remains fragile under severe degradation, often leading to identity inconsistency or structure-texture misalignment. 
In contrast, our approach seeks to mitigate these uncertainties by explicitly optimizing the restoration trajectory via task-aware feedback and auxiliary semantic guidance.

\subsection{\texorpdfstring{\textcolor{RevisionBlue}{Text-driven Semantic Guidance}}{Text-driven Semantic Guidance}}
{\color{RevisionBlue}
Text has become an effective high-level condition for image restoration because it can express semantic attributes that are difficult to infer from degraded pixels alone~\cite{wei2025pure}. Early text-driven restoration methods use CLIP-like representations~\cite{radford2021clip} or prompt-based modules to align degraded visual features with linguistic guidance. Representative works such as \textit{PromptIR}~\cite{potlapalli2023promptir}, \textit{InstructIR}~\cite{conde2024instructir}, and \textit{DA-CLIP}~\cite{luo2023controlling} show that language prompts can improve all-in-one restoration and multi-task generalization. Recent text-conditioned restoration methods further use natural-language prompts to steer restoration toward specific semantic attributes~\cite{yang2025sodiff,qi2024spire,yu2024promptfix}.

Beyond static text conditioning, native multimodal generation models such as \textit{BLIP3o-Next}~\cite{chen2025blip3o} provide strong AR-based semantic priors. Our semantic branch builds on this line, but BFR requires a different use: semantic cues should stabilize the target identity rather than drive open-ended variation. This makes deployment robustness essential, since text may be absent, degraded-derived, or imperfect.
}

\subsection{\texorpdfstring{\textcolor{RevisionBlue}{Preference Optimization for Restoration}}{Preference Optimization for Restoration}}

Preference learning has recently become an important tool for improving perceptual quality beyond pixel-wise restoration losses. Early RL-based restoration methods, such as \textit{RL-Restore}~\cite{yu2018crafting} and \textit{Path-Restore}~\cite{yu2021path}, focused on sequential tool or path selection. With the success of RLHF~\cite{ouyang2022rlhf}, recent restoration and generation methods have moved toward perceptual preference alignment. \textit{DiffusionReward}~\cite{wu2025diffusionreward} learns reward models for dense perceptual supervision, while DPO-style methods such as \textit{DPO}~\cite{rafailov2023dpo} and \textit{DSPO}~\cite{cai2025dspo} optimize from paired preference data. GRPO-style approaches~\cite{shao2024grpo}, including \textit{IRPO}~\cite{liu2025irpo}, \textit{RealSR-R1}~\cite{qiao2025realsr-r1}, and \textit{TTPO}~\cite{li2025test}, further explore online reward-driven optimization.

For diffusion and flow-based models, \textit{DiffusionNFT}~\cite{zheng2025diffusionnft} provides an efficient forward-process optimizer and naturally fits our flow-matching backbone. Yet BFR requires preference optimization to respect identity and facial structure, not only aesthetic quality. Pref-Restore therefore builds on this interface with a face-aware restoration reward that couples perceptual preference and fidelity constraints for deterministic, identity-faithful restoration.

\section{\texorpdfstring{\textcolor{RevisionBlue}{Forward-Process Preference Optimization}}{Foundations of Forward-Process Preference Optimization}}

{\color{RevisionBlue}The under-constrained nature of BFR results in a one-to-many mapping, where a single sparse input can correspond to multiple semantically plausible but structurally inconsistent outputs. To eliminate this stochastic uncertainty and ensure deterministic restoration, it is essential to constrain the generative trajectory, guiding the model's posterior distribution toward a high-fidelity manifold that aligns with human preferences. }

To this end, we leverage the principles of DiffusionNFT~\cite{chen2025nft, zheng2025diffusionnft} as our mathematical foundation. DiffusionNFT bypasses the complexities of policy gradients on the reverse process by performing RL directly on the forward diffusion process. Instead of optimizing likelihood ratios of a discretized reverse policy, it refines the diffusion dynamics through the velocity field that parameterizes the forward process. 

Specifically, given a pretrained diffusion policy $\pi_{\mathrm{old}}$ and a scalar reward $r(x_0,c) \in [0,1]$ defined on generated samples, DiffusionNFT induces two implicit data distributions:
\begin{align}
\pi^{+}(x_0 \mid c) &\propto r(x_0,c)\,\pi_{\mathrm{old}}(x_0 \mid c), \\
\pi^{-}(x_0 \mid c) &\propto (1 - r(x_0,c))\,\pi_{\mathrm{old}}(x_0 \mid c).
\end{align}
In the context of restoration, $\pi^{+}$ represents the desired high-fidelity distribution, while $\pi^{-}$ captures failure modes such as over-smoothing or identity-inconsistent hallucinations. By explicitly modeling these negative samples, policy improvement is formulated as a contrastive refinement between positive and negative generations. Under the velocity parameterization, let $v_{\mathrm{old}}$, $v^{+}$, and $v^{-}$ denote the velocity fields associated with $\pi_{\mathrm{old}}$, $\pi^{+}$, and $\pi^{-}$, respectively. A stable policy improvement direction $\Delta(x_t,t)$ is then expressed as:
\begin{equation}
\begin{aligned}
\Delta(x_t,t) &= \alpha(x_t)\big(v^{+}(x_t,t) - v_{\mathrm{old}}(x_t,t)\big) \\
&= (1 - \alpha(x_t))\big(v_{\mathrm{old}}(x_t,t) - v^{-}(x_t,t)\big),
\end{aligned}
\end{equation}
where $\alpha(x_t) \in [0,1]$ balances the attraction toward positive restorations and the repulsion from negative artifacts. 

To achieve this without training separate models for $v^{+}$ and $v^{-}$, we optimize a single network $v_\theta$ via an implicit supervised objective:
\begin{equation}
\begin{aligned}
\mathcal{L}_{\text{NFT}}(\theta)
=
\mathbb{E}\Big[
& r\,\|v_\theta^{+}(x_t,t)-v\|_2^2 \\
&+
(1-r)\,\|v_\theta^{-}(x_t,t)-v\|_2^2
\Big],
\end{aligned}
\end{equation}
where $v$ is the ground-truth velocity, and the implicit velocities are constructed as $v_\theta^{+} = (1-\gamma)v_{\mathrm{old}} + \gamma v_\theta$ and $v_\theta^{-} = (1+\gamma)v_{\mathrm{old}} - \gamma v_\theta$. {\color{RevisionBlue}This formulation provides an efficient mechanism for reward-guided trajectory refinement, which we later instantiate with a face-aware fidelity-constrained reward for deterministic restoration.}

\section{Methodology}

\begin{figure*}[!t]
\centering
\includegraphics[width=0.97\linewidth]{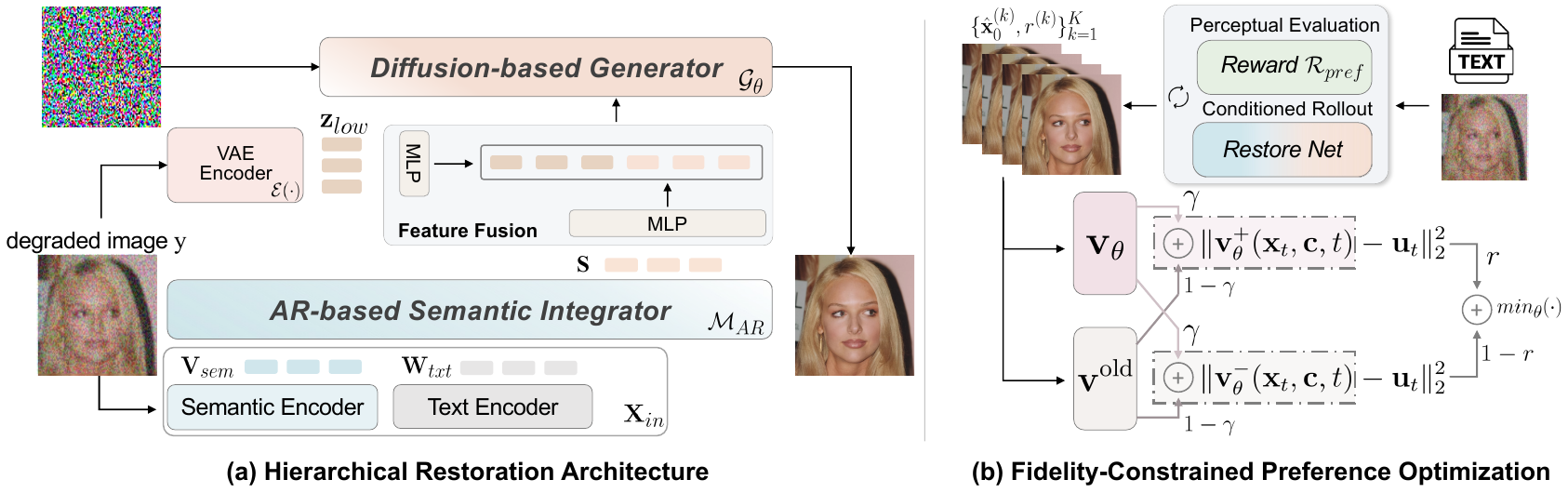}
\vspace{-1mm}
\caption{{\color{RevisionBlue}\textbf{The overall framework of Pref-Restore.} \textbf{(a) Hierarchical Restoration Architecture:} Pref-Restore decomposes blind face restoration into a semantic-guided identity branch and a continuous texture restoration branch. The AR semantic branch processes degraded observations $y$ and textual instructions $T$ to generate semantic tokens $\mathbf{S}$ via Next-Token Prediction (NTP), providing an identity-oriented anchor. The diffusion generator combines $\mathbf{S}$ with low-level texture latents $\mathbf{z}_{low}$ and reconstructs high-fidelity details through conditional flow matching. \textbf{(b) Fidelity-Constrained Preference Optimization:} The current policy $\mathbf{v}_\theta$ performs group rollout to generate $K$ candidates, which are evaluated by a fidelity-constrained reward $\mathcal{R}_{pref}$. Based on normalized rewards $r$, we construct implicit positive ($\mathbf{v}_\theta^+$) and negative ($\mathbf{v}_\theta^-$) velocity proxies and optimize the model by contrasting them against the forward data flow.}}
\vspace{-2mm}
\label{arch}
\end{figure*}

{\color{RevisionBlue}
To address the under-constrained nature of blind face restoration, we propose Pref-Restore, a hierarchical framework for deterministic, identity-faithful restoration. Pref-Restore constrains the restoration process from two complementary directions: semantic-guided identity stabilization before texture synthesis, and fidelity-constrained preference optimization during diffusion refinement. In the following sections, we first present the constrained restoration formulation (Sec.~\ref{sec:overview}), followed by the hierarchical architecture (Sec.~\ref{sec:arch}) and the staged training strategy (Sec.~\ref{sec:training}).
}

\subsection{\texorpdfstring{\textcolor{RevisionBlue}{Overview: Constrained Restoration Formulation}}{Overview: Constrained Restoration Formulation}}
\label{sec:overview}

Conventional Blind Image Restoration is formulated as the estimation of a HQ image $x \in \mathcal{X}$ from a degraded observation $y = \mathcal{D}(x) + \epsilon$, where $\mathcal{D}(\cdot)$ denotes an unknown degradation function. 
From a Bayesian perspective, this is a Maximum A Posteriori (MAP) estimation problem:
\begin{equation}
\begin{aligned}
        \hat{x}_{MAP} &= \operatorname*{arg\,max}_{x} \log p(x|y) \\
        &= \operatorname*{arg\,max}_{x} \left( \log p(y|x) + \log p(x) \right),
\end{aligned}
\end{equation}

{\color{RevisionBlue}
where $p(y|x)$ is the likelihood term representing data fidelity, and $p(x)$ is the prior term capturing natural image statistics. Because severe degradation removes identity-defining high-frequency cues, the posterior $p(x|y)$ is highly under-constrained: many plausible faces can explain the same LQ observation, but only a subset preserves the target identity.
}

{\color{RevisionBlue}
Relying solely on a generic generative prior $p(x)$ is therefore insufficient: the restoration should be (1) conditioned on identity-relevant semantics and (2) guided by a preference signal that preserves fidelity. Let $\mathcal{S}_{AR}$ denote the semantic anchor produced by the AR branch. Pref-Restore captures these two constraints by maximizing the following joint objective:
}
\begin{equation}
    \hat{x} = \operatorname*{arg\,max}_{x} \left( \underbrace{\log p(x | y, \mathcal{S}_{AR})}_{\text{Augmented Likelihood}} + \lambda \cdot \underbrace{\mathcal{R}_{pref}(x)}_{\text{Preference Constraint}} \right),
    \label{eq:objective}
\end{equation}

{\color{RevisionBlue}
where $\lambda$ is a balancing coefficient. The preference term $\mathcal{R}_{\text{pref}}(x)$ can be interpreted as an energy-based prior~\cite{lecun2006tutorial}, $p_{\text{pref}}(x) \propto \exp(\mathcal{R}_{\text{pref}}(x))$, which re-weights candidate restorations toward perceptually preferred and fidelity-preserving outputs.
}

{\color{RevisionBlue}
The two terms in Eq.~\eqref{eq:objective} correspond to two complementary constraints:
\begin{itemize}
    \item \textbf{Augmented Likelihood via Semantic-Guided Identity Stabilization}: The AR branch integrates degraded observations $y$ and textual semantics $T$ to produce the semantic anchor $\mathcal{S}_{AR}$. This anchor conditions the likelihood term with identity-relevant cues before texture synthesis, reducing drift toward plausible but incorrect faces.
    
    \item \textbf{Preference Constraint via Fidelity-Constrained Optimization}: The preference term $\mathcal{R}_{\text{pref}}(x)$ guides the diffusion trajectory toward outputs that are perceptually preferred while preserving facial fidelity. In Stage~2, this reward is instantiated with aesthetic and fidelity-related terms, making the quality--fidelity behavior controllable.
\end{itemize}

This formulation separates the two constraints needed for deterministic BFR: semantic anchoring reduces identity ambiguity, while fidelity-aware preference optimization controls the restoration trajectory.
}

\subsection{\texorpdfstring{\textcolor{RevisionBlue}{Hierarchical Architecture: Bridging Semantic Anchoring and Texture Restoration}}{Hierarchical Architecture: Semantic Anchoring and Texture Restoration}}
\label{sec:arch}

{\color{RevisionBlue}
The core of Pref-Restore is a hierarchical architecture that separates semantic-guided identity stabilization from continuous texture restoration. As illustrated in Fig.~\ref{arch} (a), the framework comprises two synergistic components: an AR-based semantic branch for high-level identity cues and a continuous diffusion generator for high-fidelity texture reconstruction. This design lets the semantic stream provide an identity-oriented anchor, while the diffusion stream restores fine-grained facial textures under that anchor.
}

\subsubsection{\textbf{AR-based Semantic Integrator}}
{\color{RevisionBlue}The AR module stabilizes identity-relevant semantics by operating in a discrete token space.} Given a degraded image $y \in \mathbb{R}^{H \times W \times 3}$ and its corresponding textual description $T$, we first extract their respective multimodal features. Specifically, a semantic encoder (e.g., Siglip2), denoted as $\Phi_{sem}(\cdot)$, projects the degraded image into a continuous visual embedding space $\mathbf{V}_{sem} = \Phi_{sem}(y) \in \mathbb{R}^{L \times C}$, where $L$ and $C$ denote the sequence length and embedding dimension, respectively. Simultaneously, the textual instruction $T$ is transformed into text embeddings $\mathbf{W}_{txt}$ via a text encoder.

The input to the AR model (e.g., Qwen3) is formed by the concatenation of these multimodal representations: $\mathbf{X}_{in} = [\mathbf{V}_{sem}; \mathbf{W}_{txt}]$. To bridge the gap between continuous signals and discrete logic, the AR module $\mathcal{M}_{AR}$ models the restoration task through a NTP paradigm over a learned visual vocabulary $\mathcal{V}$. Formally, it generates a sequence of discrete semantic tokens $\mathbf{S} = \{s_1, s_2, \dots, s_n\}$, where each token $s_i \in \mathcal{V}$ is sampled from the predicted categorical distribution:
\begin{equation}
    P(s_i \mid s_{<i}, \mathbf{X}_{in}) = \text{Softmax}(\mathcal{M}_{AR}(s_{<i}, \mathbf{X}_{in})).
\end{equation}
These discrete tokens $\mathbf{S}$ serve as a ``semantic anchor," encapsulating high-level attributes (e.g., facial identity, structural layout) that remain invariant to pixel-level degradations.

\subsubsection{\textbf{Continuous Diffusion-based Generator}}
While the AR module provides global logic, the Diffusion module $\mathcal{G}_{\theta}$ translates these priors into fine-grained textures. To compensate for the potential loss of fine-grained structures in discrete tokens, we explicitly incorporate low-level texture cues $\mathbf{z}_{low} = \mathcal{E}(y)$, where $\mathcal{E}(\cdot)$ is a pre-trained VAE encoder.

The restoration is modeled as a Conditional Flow Matching process. We seek to learn a velocity field $v_\theta$ that defines a probability path between the noise distribution and the high-fidelity latent manifold. For any timestep $t \in [0, 1]$, the generative trajectory is defined by the ODE:
\begin{equation}
    d \mathbf{z}_t = v_{\theta}(\mathbf{z}_t, t, \mathbf{S}, \mathbf{z}_{low})  d t.
\end{equation}
The DiT architecture estimates $v_{\theta}$ by synergizing the global semantic ``anchor" $\mathbf{S}$ and the local structural ``prior" $\mathbf{z}_{low}$.

\subsubsection{\textbf{Structural Synergies}}

{\color{RevisionBlue}
By decomposing the problem into discrete semantic tokens $\mathbf{S}$ and continuous texture latents $\mathbf{z}_{low}$, Pref-Restore separates identity-oriented semantic anchoring from texture-level reconstruction. The discrete stream supplies high-level identity and structure cues, while the continuous stream ensures that the final output $\hat{x} = \mathcal{D}_{vae}(\mathbf{z}_0)$ retains the perceptual granularity required for high-fidelity restoration.
}

\subsection{\texorpdfstring{\textcolor{RevisionBlue}{Training Strategy: Staged Alignment and Fidelity-Guided Preference Refinement}}{Training Strategy: Staged Alignment and Fidelity-Guided Preference Refinement}}
\label{sec:training}

{\color{RevisionBlue}
Pref-Restore is trained with a progressive two-stage strategy that first aligns semantic and texture conditions with the diffusion generator, and then refines the restoration trajectory through fidelity-constrained preference optimization. \textit{Multi-modal Knowledge Alignment} maps AR-derived semantic tokens and VAE-encoded texture cues into the diffusion latent space, enabling identity-aware guidance during texture synthesis. \textit{Preference-Aware Fine-tuning} further optimizes the flow trajectory with a face-aware reward, improving perceptual quality while maintaining identity and structural fidelity.
}

\subsubsection{\textbf{Stage 1: Multi-modal Knowledge Alignment}}

{\color{RevisionBlue}
The first training stage establishes a shared conditioning space between discrete semantic reasoning and continuous image reconstruction. We adopt a staged design because jointly updating the AR semantic branch and the diffusion texture generator makes both the conditioning signal and reconstruction target move simultaneously, which can destabilize identity anchoring. We therefore first align semantic tokens to the diffusion attention space, and then train the texture path under this fixed semantic anchor. This process is divided into two objectives: \textit{Semantic-to-Diffusion Alignment} and \textit{Texture-to-Diffusion Alignment}.
}

{\color{RevisionBlue}
\textbf{Stage 1.1 Semantic-to-Diffusion Alignment. }
Building on the AR semantic branch, this stage focuses on making its token outputs compatible with the diffusion generator for BFR-specific identity guidance. We freeze the diffusion generator $\mathcal{G}_{\theta}$ and optimize $\mathcal{M}_{AR}$ together with a cross-modal projector $\mathcal{P}$, so that the generated tokens can serve as conditions for the frozen diffusion generator. The alignment loss is:
}
\begin{equation}
    \mathcal{L}_{align} = \mathcal{L}_{CE} + \alpha \mathcal{L}_{diff},
\end{equation}

{\color{RevisionBlue}
where $\mathcal{L}_{CE}$ is the cross-entropy loss over the expanded visual vocabulary $\mathcal{V}_{img} = \{\langle I_0 \rangle, \dots, \langle I_{65535} \rangle\}$, and $\mathcal{L}_{diff}$ is diffusion denoising loss. Gradients from $\mathcal{L}_{diff}$ are backpropagated through the frozen $\mathcal{G}_{\theta}$ to $\mathcal{P}$ and $\mathcal{M}_{AR}$. In this way, the semantic tokens $\mathbf{S}$ remain supervised by token prediction while compatible with the diffusion module's conditioning space.
}

{\color{RevisionBlue}
\textbf{Stage 1.2 Texture-to-Diffusion Alignment. }
After the semantic anchor is aligned, we freeze $\mathcal{M}_{AR}$ and fine-tune the VAE encoder $\mathcal{E}$ together with the diffusion generator $\mathcal{G}_{\theta}$. This sub-stage makes $\mathbf{z}_{low}=\mathcal{E}(y)$ a reliable low-level condition for preserving identity-sensitive structure and local texture cues. We introduce a latent-level consistency loss:
}
\begin{equation}
    \mathcal{L}_{mse} = \| \mathcal{E}(y) - \mathcal{E}(x) \|^2_2,
\end{equation}

{\color{RevisionBlue}
where $y$ and $x$ denote the degraded and ground-truth images, respectively. Together with the diffusion loss, this consistency term encourages the low-level path to retain information shared by degraded and clean faces, while $\mathcal{G}_{\theta}$ learns to synthesize details conditioned on both $\mathbf{S}$ and $\mathbf{z}_{low}$. The total objective for this stage is $\mathcal{L}_{stage1} = \mathcal{L}_{diff} + \beta \mathcal{L}_{mse}$.
}

\subsubsection{\textbf{\textcolor{RevisionBlue}{Stage 2: Fidelity-Constrained Preference Optimization}}}
{\color{RevisionBlue}
After Stage~1 establishes semantic and texture conditioning, the diffusion trajectory still needs face-specific preference guidance to avoid visually plausible but fidelity-weak details. We therefore build Stage~2 on the forward-process optimizer of DiffusionNFT~\cite{zheng2025diffusionnft}, and introduce a BFR-specific reward that couples perceptual preference with identity, similarity, and geometry constraints. This makes preference optimization a fidelity-constrained refinement mechanism for face restoration.

\textbf{Fidelity-Constrained Reward. }
For a restored candidate $\hat{x}$ and its reference image $x$, we define the preference reward:
}
\begin{equation}
\begin{aligned}
    \mathcal{R}_{\text{pref}}(\hat{x}, x) =
    & \lambda_{\text{aes}} \mathcal{R}_{\text{aes}}(\hat{x})
    + \lambda_{\text{id}} \mathcal{R}_{\text{id}}(\hat{x}, x) \\
    & + \lambda_{\text{lpips}} \mathcal{R}_{\text{lpips}}(\hat{x}, x)
    + \lambda_{\text{geo}} \mathcal{R}_{\text{geo}}(\hat{x}, x),
    \label{eq:pref_reward}
\end{aligned}
\end{equation}
{\color{RevisionBlue}
where $\mathcal{R}_{\text{aes}}$ aggregates preference scores such as HPSv2~\cite{wu2023hpsv2}, CLIPScore~\cite{hessel2021clipscore}, and PickScore~\cite{kirstain2023pick}; $\mathcal{R}_{\text{id}}$ measures ArcFace~\cite{deng2019arcface} identity similarity; $\mathcal{R}_{\text{lpips}}$ is a LPIPS-based similarity reward~\cite{zhang2018lpips}; and $\mathcal{R}_{\text{geo}}$ measures landmark consistency. All terms are normalized before weighted summation.

\textbf{Conditioned Rollout and Reward Evaluation. }
As outlined in Fig.~\ref{arch} (b), for a given condition $\mathbf{c} = (\mathbf{S}, \mathbf{z}_{low})$, the current policy $\pi_{\theta}$ samples $K$ candidate latent restorations $\{\hat{\mathbf{x}}_0^{(k)}\}_{k=1}^K$. After decoding each candidate to image space, we evaluate it with $\mathcal{R}_{\text{pref}}$ and obtain raw rewards $r^\text{raw}_{(k)}$. To reduce reward-scale variation across samples, we convert raw rewards into normalized optimality probabilities $r^{(k)} \in [0, 1]$ through group centering and clipping:
}
\begin{equation}
    r^{(k)} = 0.5 + 0.5 \cdot \operatorname{clip}\left(\frac{r^\text{raw}_{(k)} - \mu_{group}}{Z}, -1, 1\right),
\end{equation}
{\color{RevisionBlue}
where $\mu_{group}$ is the mean reward of the rollout group, and $Z$ is a scaling factor. A larger $r^{(k)}$ indicates a candidate more favored by the fidelity-constrained reward, while a smaller $r^{(k)}$ indicates a lower-reward candidate under the same condition.

\textbf{Implicit Velocity Proxies. } 
Following DiffusionNFT, we avoid back-propagating through the full ODE solver by constructing implicit positive ($\mathbf{v}_\theta^+$) and negative ($\mathbf{v}_\theta^-$) velocity proxies. Let $\mathbf{v}^\text{old}$ denote the Stage~1 reference vector field and $\mathbf{v}_\theta$ denote the current policy. The proxies are defined as:
}
\begin{align}
    \mathbf{v}_\theta^+(\mathbf{x}_t) &:= (1-\gamma) \mathbf{v}^\text{old}(\mathbf{x}_t, \mathbf{c}, t) + \gamma \mathbf{v}_\theta(\mathbf{x}_t, \mathbf{c}, t), \\
    \mathbf{v}_\theta^-(\mathbf{x}_t) &:= (1+\gamma) \mathbf{v}^\text{old}(\mathbf{x}_t, \mathbf{c}, t) - \gamma \mathbf{v}_\theta(\mathbf{x}_t, \mathbf{c}, t).
\end{align}
{\color{RevisionBlue}
The positive proxy reinforces trajectory deviations associated with higher reward under $\mathcal{R}_{\text{pref}}$, while the negative proxy regularizes lower-reward deviations.

\textbf{Preference-Weighted Flow Matching. }
The model is optimized by a reward-weighted contrastive flow matching objective:
}
\begin{equation}
\begin{aligned}
    \mathcal{L}_{RL} = \mathbb{E}_{\substack{t \sim \mathcal{U}(0,1) \\ \hat{\mathbf{x}}_0 \sim \pi_{\theta}, \mathbf{c}}} \Big[ & r \| \mathbf{v}_\theta^+ (\mathbf{x}_t, \mathbf{c}, t) - \mathbf{u}_{t} \|_2^2 \\
    & + (1-r) \| \mathbf{v}_\theta^- (\mathbf{x}_t, \mathbf{c}, t) - \mathbf{u}_{t} \|_2^2 \Big],
    \label{eq:nft_loss}
\end{aligned}
\end{equation}
{\color{RevisionBlue}
where $\mathbf{u}_t$ is the target forward velocity. The reward weight $r$ increases the influence of high-reward trajectories, while the $1-r$ term suppresses lower-reward directions. This refines the diffusion vector field toward a controllable quality--fidelity operating point for face restoration.
}


\section{Experiments}

\subsection{Experimental Settings}

\textbf{Implementation Details.} 
We initialize our Pref-Restore with the weights of \textit{blip-3o-next}\cite{chen2025blip3o} to leverage its pre-aligned multi-modal features. 
In the AR-based semantic integrator, we employ \textit{Siglip2}\cite{tschannen2025siglip} as the semantic encoder $\Phi_{sem}$. The degraded images are resized to $384 \times 384$ before being fed into a lightweight \textit{Qwen3-0.6B} backbone. During the NTP process, the model learns to predict 729 tokens from a visual vocabulary $\mathcal{V}_{img}$ of size 65,536, which serve as high-level semantic anchors. For the generative component, we adopt the \textit{SANA-1.5\_1.6B}\cite{xie2025sana} as its efficiency.

{\color{RevisionBlue}
To adapt the \textit{blip-3o-next}-based framework to BFR, we train Pref-Restore through a three-step pipeline (Sec.~\ref{sec:training}): (1) Stage~1.1: Semantic-to-Diffusion Alignment; (2) Stage~1.2: Texture-to-Diffusion Alignment; and (3) Stage~2: Fidelity-Constrained Preference Optimization with the reward defined in Eq.~\eqref{eq:pref_reward}. Training hyperparameters and schedules are summarized in Table~\ref{tab:training_recipe}. All experiments were implemented using PyTorch and conducted with 8 NVIDIA A100 GPUs.
}

\begin{table}[!t]
    \caption{Training recipe of Pref-Restore.}
    \label{tab:training_recipe}
    \vspace{-2mm}
    \centering
    \resizebox{0.47\textwidth}{!}{
    \begin{tabular}{l|ccc}
        \toprule
         \textbf{Hyperparameters} & Stage-1.1 (Step1) & Stage-1.2 (Step2) & Stage-2  \\
         \midrule
         Learning rate & $1\times 10^{-5}$ & $1\times 10^{-5}$ & $3\times 10^{-4}$ \\ 
         LR scheduler & Cosine & Cosine  & Constant\\
         Weight decay & 0.0 & 0.0 & $1\times 10^{-4}$  \\
         Optimizer & \multicolumn{3}{c}{AdamW ($\beta_1 = 0.9, \beta_2 = 0.999, \epsilon = 1 \times 10^{-8}$)} \\
         Loss type & CE+MSE(diff) & MSE(VAE+diff) & MSE(diff) \\
         Warm-up ratio & 0.01 & 0.01 & 0 \\
         Batch size & 128 & 128 & 72 \\
         Module & Qwen3 & SANA+VAE & SANA\\
         
        \bottomrule
    \end{tabular}
    }
    \vspace{-3mm}
\end{table}

\textbf{Datasets.} 
Following the common practice in blind face restoration \cite{yang2021gpen, zhou2022codeformer, wang2021gfpgan}, we use the FFHQ dataset \cite{karras2019style} (70,000 high-quality images) for training. All images are resized to $512 \times 512$. To synthesize degraded images, we utilize the following degradation model:
\begin{equation}\label{LQgenerator}
I_l = \{ [(I_h \otimes k_{\sigma})\downarrow_{r} + n_{\delta}]_{\mathrm {JPEG}_q} \}\uparrow_r,
\end{equation}
where $\sigma \in [1, 15]$, $r \in [1, 30]$, $\delta \in [0, 20]$, and $q \in [40, 100]$ are randomly sampled to simulate complex real-world artifacts.

For evaluation, we utilize one synthetic and four real-world datasets:

\begin{itemize}
    \item CelebA-Test: A synthetic dataset comprising 3,000 images from CelebA-HQ \cite{karras2017celebA}, with LQ versions generated via Eq. \eqref{LQgenerator}.
    \item LFW-Test \cite{huang2008lfw}: 1,711 mildly degraded real-world images (one per individual).
    \item WIDER-Test \cite{yang2016wider}: 970 heavily degraded real-world images from the WIDER Face dataset.
    \item WebPhoto-Test \cite{wang2021gfpgan}: 407 real-world face images collected from the internet, including historical photos that exhibit complex and severe degradations.
    \item CelebChild-Test \cite{wang2021gfpgan}: 180 child faces of celebrities, exhibiting unique degradation and cross-age patterns.
\end{itemize}

\textbf{Evaluation Metrics.} 
To provide comprehensive evaluations, we categorize our metrics into general image quality and face-specific fidelity assessment. 
For general quality, we employ reference-based metrics including LPIPS \cite{zhang2018lpips} and FID \cite{heusel2017fid}, alongside no-reference metrics for perceptual quality such as MUSIQ\cite{ke2021musiq}, CLIPIQA+\cite{wang2023clipiqa}, and MANIQA\cite{yang2022maniqa}. \footnote{We exclude PSNR and SSIM, as they tend to penalize high-frequency generative details in severely degraded scenarios, often yielding misleadingly high scores for over-smoothed results.}

For face-specific fidelity, we adopt the ArcFace embedding angle (Deg) and Landmark Distance (LMD) \cite{deng2019arcface} to evaluate identity preservation and structural accuracy. Additionally, we use topiq\_nr-face, topiq\_nr\_swin-face, and DSL-FIQA as specialized no-reference metrics to quantify the aesthetic and biological plausibility of the restored faces. \footnote{These metrics are implemented based on the IQA-PyTorch toolbox \cite{pyiqa}.}

\subsection{Quantitative Comparison}

{\color{RevisionBlue}
\textbf{Results on Synthetic Datasets.} 
We conduct a comprehensive comparison between our Pref-Restore and several SOTA methods, including diffusion-based approaches (DR2 \cite{wang2023dr2}, DifFace \cite{yue2024difface}, and DiffBIR \cite{lin2023diffbir}), GAN-based models (GFPGAN \cite{wang2021gfpgan}, GPEN \cite{yang2021gpen}), and VQ-based techniques (CodeFormer \cite{zhou2022codeformer}, VQFR \cite{gu2022vqfr}, RestoreFormer++ \cite{wang2023restoreformer++}, and DAEFR\cite{tsai2023DAEFR}).

As summarized in Table~\ref{tab:celebA}, we report Pref-Restore after Stage~1 semantic--texture alignment and Pref-Restore (RL) after Stage~2 fidelity-constrained preference optimization. This comparison shows how the hierarchical framework first establishes a strong restoration base and then refines the quality--fidelity metrics through reward-guided trajectory optimization.

Regarding reconstruction fidelity, Pref-Restore achieves the best distribution alignment, with FID(HQ) $13.9015$ and FID(FFHQ) $39.2214$, while also obtaining the second-best LPIPS ($0.3916$) and ArcFace Deg ($54.9791$). These results indicate that the Stage~1 semantic--texture alignment provides a stable identity-oriented restoration base, especially in full-reference and face-sensitive metrics.

Stage~2 further improves most perceptual and fidelity indicators. Compared with Pref-Restore, Pref-Restore (RL) improves LPIPS from $0.3916$ to $0.3657$, LMD from $5.0915$ to $4.1085$, and ArcFace Deg from $54.9791$ to $51.5394$, while also increasing CLIPIQA+ from $0.6107$ to $0.6778$, topiq from $0.7135$ to $0.8144$, and topiq-Swin from $0.8034$ to $0.8630$. This supports the role of the fidelity-constrained reward: Stage~2 does not merely enhance visual preference, but also strengthens identity- and structure-related restoration fidelity.

DiffBIR obtains very strong no-reference aesthetic scores on CelebA-Test, even exceeding the FFHQ training-data row in several such metrics. This suggests that synthetic no-reference scores can be influenced by the aesthetic distribution of training data and the matched degradation setting. However, in the real-world benchmarks (Tables \ref{tab:lfw} - \ref{tab:celebchild}), this advantage does not transfer consistently under in-the-wild degradations, where Pref-Restore achieves stronger blind robustness across most perceptual and face-plausibility metrics.
}

\begin{table*}[!ht]
\caption{{\color{RevisionBlue}\textbf{Quantitative comparison on the synthetic CelebA-Test dataset.} \textbf{Pref-Restore} denotes the Stage~1 model after semantic--texture alignment, while \textbf{Pref-Restore (RL)} denotes the Stage~2 model refined by fidelity-constrained preference optimization. \red{Red} and \blue{blue} indicate the best and second-best performance, respectively.}}
    \label{tab:celebA}
    \centering
    \resizebox{\textwidth}{!}{
    \begin{tabular}{l|cccccc|ccccc}
        \toprule
        \hfill \textbf{CelebA-Test} & \multicolumn{6}{c|}{\textbf{General Image Quality Evaluation}} & \multicolumn{5}{c}{\textbf{Face Image Quality Evaluation}} \\ 
        Methods & LPIPS$\downarrow$  & FID(HQ)$\downarrow$  & FID(FFHQ)$\downarrow$ & MUSIQ$\uparrow$ & CLIPIQA+$\uparrow$ & MANIQA$\uparrow$ & LMD$\downarrow$ & Deg$\downarrow$ & topiq$\uparrow$ & topiq-swin$\uparrow$ & DSL-FIQA$\uparrow$ \\ 
        \midrule
        FFHQ (Training Data) & - & 39.9988 & -       & 74.0405 & 0.6397 & 0.5033 & -      & -        & 0.7266 & 0.8075 & 0.7976 \\
        CelebA-Test (GT)     & -      &-        & 39.9988 & 72.8117 & 0.6218 & 0.4966 & -      & -        & 0.6778 & 0.7766 & 0.7447 \\
        \midrule
        GFPGAN          & 0.4315 & 21.7414 & 53.4286 & 75.3721 & 0.6529 & 0.5737 & 5.1382 & 58.0687 & 0.8018 & 0.8559 & \blue{0.8838}  \\  
        GPEN            & 0.4359 & 33.2368 & 49.9732 & 74.9878 & 0.6031 & 0.4375 & 5.9907 & 59.8849 & 0.668 & 0.7927 & 0.7275  \\  
        DR2             & 0.4599 & 48.7970 & 76.1381 & 63.8542 & 0.5233 & 0.5021 & 6.5405 & 66.7169 & 0.6507 & 0.7348 & 0.7108  \\  
        DifFace         & 0.4389 & \blue{18.7722} & 47.3185 & 66.7451 & 0.5570 & 0.4975 & 5.4759 & 63.4511 & 0.6730 & 0.7560 & 0.7581  \\  
        RestoreFormer++ & 0.4763 & 45.0251 & \blue{43.0445} & 72.3612 & 0.6220 & 0.5687 & 8.8019 & 70.5083 & 0.7319 & 0.7955 & 0.8301  \\  
        DAEFR           & 0.4317 & 20.9941 & 48.0317 & 74.5893 & 0.6738 & \blue{0.5761} & 5.6332 & 62.8184 & 0.7851 & 0.8380 & 0.8580  \\  
        Codeformer      & 0.4005 & 27.4670 & 65.4874 & \blue{75.9432} & 0.6584 & 0.5527 & 5.3819 & 62.1972 & 0.7946 & 0.8436 & 0.8688  \\  
        VQFR            & 0.4095 & 20.5485 & 54.2846 & 73.2969 & 0.6316 & 0.5531 & 5.7297 & 62.2595 & 0.7764 & 0.8239 & 0.8567  \\  
        DiffBIR         & 0.4594 & 25.9525 & 65.5940 & \red{76.0005} & \red{0.7420} & \red{0.7194} & \blue{4.3139} & 55.1818 & \red{0.9007} & \red{0.9229} & \red{0.9486} \\
        \rowcolor{lightblue}
        \textbf{Pref-Restore} & \blue{0.3916} & \red{13.9015} & \red{39.2214} & 72.6841 & 0.6107 & 0.4755 & 5.0915 & \blue{54.9791} & 0.7135 & 0.8034 & 0.7730  \\ 
        \rowcolor{lightblue}
        \textbf{Pref-Restore (RL)} & \red{0.3657} & 23.7305 & 66.2883 & 74.1459 & \blue{0.6778} & 0.5384 & \red{4.1085} & \red{51.5394} & \blue{0.8144} & \blue{0.8630} & 0.8176 \\  
        \bottomrule
    \end{tabular}
    }
\end{table*}

\begin{figure*}[!t]
\centering
\includegraphics[width=0.9\linewidth]{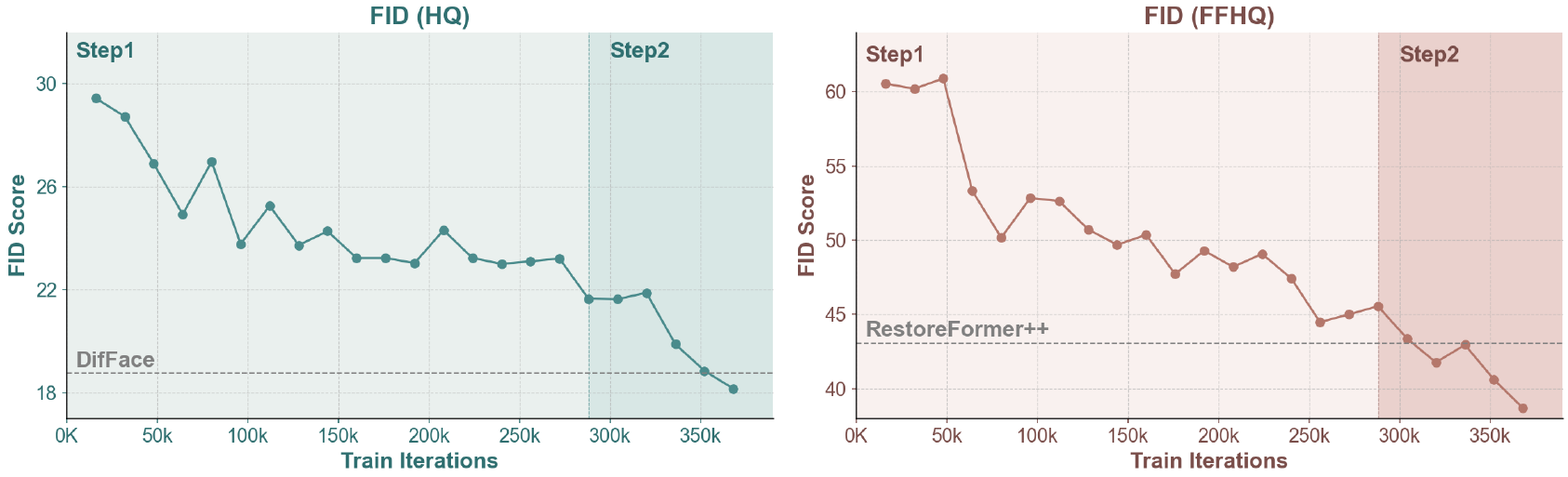}
\vspace{-3mm}
\caption{{\color{RevisionBlue}\textbf{Evolution of FID scores during the hierarchical training process.} 
    The curves illustrate FID(HQ) and FID(FFHQ) on CelebA-Test. 
    Stage~1 contains Semantic-to-Diffusion Alignment and Texture-to-Diffusion Alignment. 
    The steady decrease in FID indicates that staged semantic--texture alignment improves distribution-level restoration quality before preference refinement.}}
    \label{fig:fid_curve}
\vspace{-3mm}
\end{figure*}

{\color{RevisionBlue}
\textbf{Discussion on the Quality--Fidelity Operating Point.} 
The results in Table~\ref{tab:celebA} show that Stage~2 changes the operating point of the restoration model rather than acting as a simple aesthetic-only enhancement. While the FID scores of Pref-Restore (RL) become less favorable, the fidelity-constrained reward improves LPIPS, LMD, and ArcFace Deg together with several perceptual face-quality metrics. This behavior is consistent with our goal of making preference optimization controllable for BFR, with detailed reward-composition and operating-point analyses provided in the ablation section.
}

%
{\color{RevisionBlue}
\textbf{Scaling Analysis and Distribution Alignment.} 
As shown in Fig.~\ref{fig:fid_curve}, FID decreases consistently through Stage~1: \textit{Semantic-to-Diffusion Alignment} first maps discrete semantic tokens into the generative latent space, and \textit{Texture-to-Diffusion Alignment} further incorporates fine-grained priors $\mathbf{z}_{low}$. By the end of Stage~1, Pref-Restore achieves the best FID(HQ) and FID(FFHQ) among restored methods in Table~\ref{tab:celebA}, supporting the role of staged alignment in distribution-level restoration quality.
}

{\color{RevisionBlue}
\textbf{Results on Real-World Datasets.} 
To further validate the generalization and robustness of Pref-Restore, we conduct evaluations on four representative real-world datasets: LFW-Test, WebPhoto-Test, WIDER-Test, and CelebChild-Test. These datasets contain diverse degradations that differ from the synthetic training distribution. The quantitative results are summarized in Tables~\ref{tab:lfw}--\ref{tab:celebchild}.
}

{\color{RevisionBlue}
The real-world results show a more nuanced but stronger robustness pattern than the synthetic benchmark alone. On the mildly degraded LFW-Test, DiffBIR attains the best scores on several no-reference metrics, while Pref-Restore remains competitive and achieves the best topiq-Swin. On the more challenging WebPhoto-Test, WIDER-Test, and CelebChild-Test, Pref-Restore ranks first on most perceptual and face-plausibility metrics, including MUSIQ, topiq, topiq-Swin, and DSL-FIQA on all three datasets.

This trend suggests that the strong synthetic no-reference performance of DiffBIR does not consistently transfer to in-the-wild degradations. In particular, Pref-Restore performs best on WIDER-Test in FID, MUSIQ, topiq, topiq-Swin, and DSL-FIQA, and achieves the best scores on six of seven metrics on CelebChild-Test. Since these datasets lack ground-truth references, the metrics should be interpreted as indicators of perceptual quality and blind face plausibility rather than strict identity fidelity. Under this evaluation protocol, Pref-Restore demonstrates robust generalization across diverse real-world degradations.
}

\begin{table*}[!ht]
    \centering
    \caption{{\color{RevisionBlue}\textbf{Quantitative comparison on the real-world LFW-Test dataset.} LFW-Test contains mildly degraded real-world faces. Since no ground truth is available, these no-reference metrics reflect perceptual quality and blind face plausibility rather than strict identity fidelity. \red{Red} and \blue{blue} indicate best and second-best performance, respectively.}}
    \vspace{-2mm}
\label{tab:lfw}     
    \resizebox{0.74\textwidth}{!}{
    \begin{tabular}{l|cccc|ccc}
        \toprule
        \hfill \textbf{LFW-Test} & \multicolumn{4}{c|}{\textbf{Natural Image Quality Evaluation}} & \multicolumn{3}{c}{\textbf{Face Image Quality Evaluation}}\\ 
        Methods & FID$\downarrow$ & MUSIQ$\uparrow$ & CLIPIQA+$\uparrow$ & MANIQA$\uparrow$ & topiq$\uparrow$ & topiq-swin$\uparrow$ & DSL-FIQA$\uparrow$ \\ 
        \midrule
        GFPGAN & 48.6610  & 76.5157 & 0.6630 & 0.5843 & 0.8217 & 0.8765 & 0.8811 \\ 
        GPEN & 55.0946  & 72.3826 & 0.5946 & 0.4606 & 0.7308 & 0.8191 & 0.7863 \\ 
        DR2 & 53.5473  & 72.4804 & 0.5970 & 0.5308 & 0.7481 & 0.8156 & 0.8231 \\ 
        DifFace & \blue{47.0041}  & 69.8269 & 0.5642 & 0.4600 & 0.6727 & 0.7613 & 0.7554 \\ 
        RestoreFormer & 49.3172 & 73.6487 & 0.6160 & 0.5657 & 0.7464 & 0.8043 & 0.8325 \\ 
        DAEFR & 48.5389  & 75.8789 & 0.6559 & 0.5425 & 0.7744 & 0.8369 & 0.8365 \\ 
        Codeformer & 52.4015  & 75.4539 & 0.6480 & 0.5231 & 0.7787 & 0.8404 & 0.8431 \\ 
        VQFR & 50.9369  & 74.5095 & 0.6077 & 0.5353 & 0.7617 & 0.8257 & 0.8352 \\ 
        DiffBIR & \red{44.3631} & \red{76.8781} & \red{0.7364} & \red{0.6942} & \red{0.8707} & \blue{0.9117} & \red{0.9076} \\
        \rowcolor{lightblue}
        \textbf{Pref-Restore} & 54.7887  & \blue{76.6660} & \blue{0.6951} & \blue{0.5883} & \blue{0.8471} & \red{0.9121} & \blue{0.8913} \\ 
        \bottomrule
    \end{tabular}
    }
\end{table*}

\begin{table*}[!ht]
    \centering
    \caption{{\color{RevisionBlue}\textbf{Quantitative comparison on the real-world WebPhoto-Test dataset.} WebPhoto-Test contains complex historical and internet photos with diverse degradations. The no-reference metrics assess perceptual quality, distributional realism, and blind face plausibility under in-the-wild conditions. \red{Red} and \blue{blue} indicate best and second-best performance, respectively.}}
    \vspace{-2mm}
\label{tab:webphoto}
    \resizebox{0.74\textwidth}{!}{
    \begin{tabular}{l|cccc|ccc}
        \toprule
        \hfill \textbf{Webphoto-Test} & \multicolumn{4}{c|}{\textbf{Natural Image Quality Evaluation}} & \multicolumn{3}{c}{\textbf{Face Image Quality Evaluation}}\\ 
        Methods & FID$\downarrow$ & MUSIQ$\uparrow$ & CLIPIQA+$\uparrow$ & MANIQA$\uparrow$ & topiq$\uparrow$ & topiq-swin$\uparrow$ & DSL-FIQA$\uparrow$ \\ 
        \midrule
        GFPGAN & 91.7843 & \blue{74.7166} & 0.6157 & 0.5366 & \blue{0.7622} & \blue{0.8146} & 0.8152  \\ 
        GPEN & 86.5962 & 65.6675 & 0.4886 & 0.4290 & 0.6466 & 0.7339 & 0.7094  \\ 
        DR2 & 123.5743 & 64.1698 & 0.5038 & 0.4510 & 0.6160 & 0.6946 & 0.6669  \\ 
        DifFace & 82.6670 & 65.5357 & 0.4940 & 0.4236 & 0.5924 & 0.6900 & 0.6647  \\ 
        RestoreFormer & \blue{79.6334} & 69.8371 & 0.5430 & 0.5183 & 0.6653 & 0.7337 & 0.7267  \\ 
        DAEFR & \red{77.4463} & 72.7081 & 0.5944 & 0.5140 & 0.7029 & 0.7695 & 0.7705  \\ 
        Codeformer & 92.6693 & 71.9399 & 0.5930 & 0.4831 & 0.7079 & 0.7665 & \blue{0.8279}  \\
        VQFR & 88.3393 & 70.9087 & 0.5675 & 0.4909 & 0.6994 & 0.7534 & 0.7601  \\ 
        DiffBIR & 94.2471 & 72.8021 & \blue{0.6607} & \red{0.6038} & 0.7576 & 0.8043 & 0.7815 \\
        \rowcolor{lightblue}
        \textbf{Pref-Restore} & 85.5779 & \red{76.0059} & \red{0.7035} & \blue{0.5384} & \red{0.7837} & \red{0.8506} & \red{0.8486} \\ 
        \bottomrule
    \end{tabular}
    }
\end{table*}

\begin{table*}[!ht]
    \centering
    \caption{{\color{RevisionBlue}\textbf{Quantitative comparison on the real-world WIDER-Test dataset.} WIDER-Test contains heavily degraded real-world faces and evaluates blind robustness under severe artifacts. These no-reference metrics indicate perceptual quality and face plausibility, not strict identity fidelity. \red{Red} and \blue{blue} indicate best and second-best performance, respectively.}}
    \vspace{-2mm}
\label{tab:wider}
    \resizebox{0.74\textwidth}{!}{
    \begin{tabular}{l|cccc|ccc}
        \toprule
        \hfill \textbf{WIDER-Test} & \multicolumn{4}{c|}{\textbf{Natural Image Quality Evaluation}} & \multicolumn{3}{c}{\textbf{Face Image Quality Evaluation}}\\ 
        Methods & FID$\downarrow$ & MUSIQ$\uparrow$ & CLIPIQA+$\uparrow$ & MANIQA$\uparrow$ & topiq$\uparrow$ & topiq-swin$\uparrow$ & DSL-FIQA$\uparrow$ \\ 
        \midrule
        GFPGAN & 42.9806 & 74.8329 & 0.6431 & 0.5551 & 0.7949 & 0.8546 & 0.8556  \\  
        GPEN & 55.4026 & \blue{75.5878} & 0.6037 & 0.4459 & 0.7166 & 0.8224 & 0.7600  \\  
        DR2 & 54.8225 & 65.3340 & 0.5254 & 0.4751 & 0.6792 & 0.7560 & 0.7401  \\  
        DifFace & 38.4898 & 65.0191 & 0.5377 & 0.4326 & 0.6406 & 0.7367 & 0.7131  \\  
        RestoreFormer & 51.3830 & 67.8415 & 0.5644 & 0.4984 & 0.6707 & 0.7544 & 0.7532  \\  
        DAEFR & 37.6716 & 74.1470 & 0.6437 & 0.5204 & 0.7476 & 0.8151 & 0.8134  \\  
        Codeformer & \blue{37.6606} & 71.8923 & 0.6160 & 0.4816 & 0.7316 & 0.7993 & 0.7977  \\  
        VQFR & 38.8920 & 71.4168 & 0.5901 & 0.5043 & 0.7317 & 0.7977 & 0.8024  \\  
        DiffBIR & 38.4889 & 75.3087 & \red{0.7345} &  \red{0.6963} & \blue{0.8127} & \blue{0.8711} & \blue{0.8776} \\
        \rowcolor{lightblue}
        \textbf{Pref-Restore} & \red{35.1473} & \red{76.2551} & \blue{0.6913} & \blue{0.5828} & \red{0.8431} & \red{0.9020} & \red{0.8892} \\ 
        \bottomrule
    \end{tabular}
    }
\end{table*}

\begin{table*}[!ht]
    \centering
    \caption{{\color{RevisionBlue}\textbf{Quantitative comparison on the real-world CelebChild-Test dataset.} CelebChild-Test evaluates restoration on child faces with cross-age appearance and real-world degradations. The reported no-reference metrics reflect perceptual quality and face plausibility under this challenging setting. \red{Red} and \blue{blue} indicate best and second-best performance, respectively.}}
    \vspace{-2mm}
\label{tab:celebchild}
    \resizebox{0.74\textwidth}{!}{
    \begin{tabular}{l|cccc|ccc}
        \toprule
        \hfill \textbf{CelebChild-Test} & \multicolumn{4}{c|}{\textbf{Natural Image Quality Evaluation}} & \multicolumn{3}{c}{\textbf{Face Image Quality Evaluation}}\\ 
        Methods & FID$\downarrow$ & MUSIQ$\uparrow$ & CLIPIQA+$\uparrow$ & MANIQA$\uparrow$ & topiq$\uparrow$ & topiq-swin$\uparrow$ & DSL-FIQA$\uparrow$ \\ 
        \midrule
        GFPGAN & 120.9049 & \blue{74.1155} & 0.6166 & 0.5345 & 0.7484 & 0.7917 & \blue{0.8419}  \\  
        GPEN & 109.1407 & 70.3653 & 0.5491 & 0.4307 & 0.6393 & 0.7341 & 0.7429  \\  
        DR2 & 137.0317 & 69.6723 & 0.6025 & 0.4982 & 0.6918 & 0.7776 & 0.8063  \\  
        DifFace & 106.2445 & 66.6574 & 0.5285 & 0.4206 & 0.5935 & 0.7021 & 0.7079  \\  
        RestoreFormer & \red{102.8651} & 70.5183 & 0.5708 & 0.5425 & 0.6840 & 0.7474 & 0.7921  \\  
        DAEFR & 108.2202 & 73.7093 & 0.6240 & 0.5141 & 0.7189 & 0.7961 & 0.8154  \\  
        Codeformer & 124.3481 & 73.6569 & 0.6171 & 0.4985 & 0.7342 & \blue{0.8020} & 0.8279  \\  
        VQFR & 117.9200 & 72.3455 & 0.5962 & 0.5073 & 0.7090 & 0.7819 & 0.8154  \\  
        DiffBIR & 119.6238 & 73.2463 & \blue{0.6505} & \blue{0.5665} & \blue{0.7614} & 0.7955 & 0.8129 \\
        \rowcolor{lightblue}
        \textbf{Pref-Restore} & \blue{105.9776} & \red{76.0753} & \red{0.7016} & \red{0.5727} & \red{0.8121} & \red{0.8744} & \red{0.8859} \\
        \bottomrule
    \end{tabular}
    }
\end{table*}

\subsection{Qualitative Evaluation on Synthetic Datasets}

\begin{figure*}[!t]
    \centering
    \includegraphics[width=0.9\linewidth]{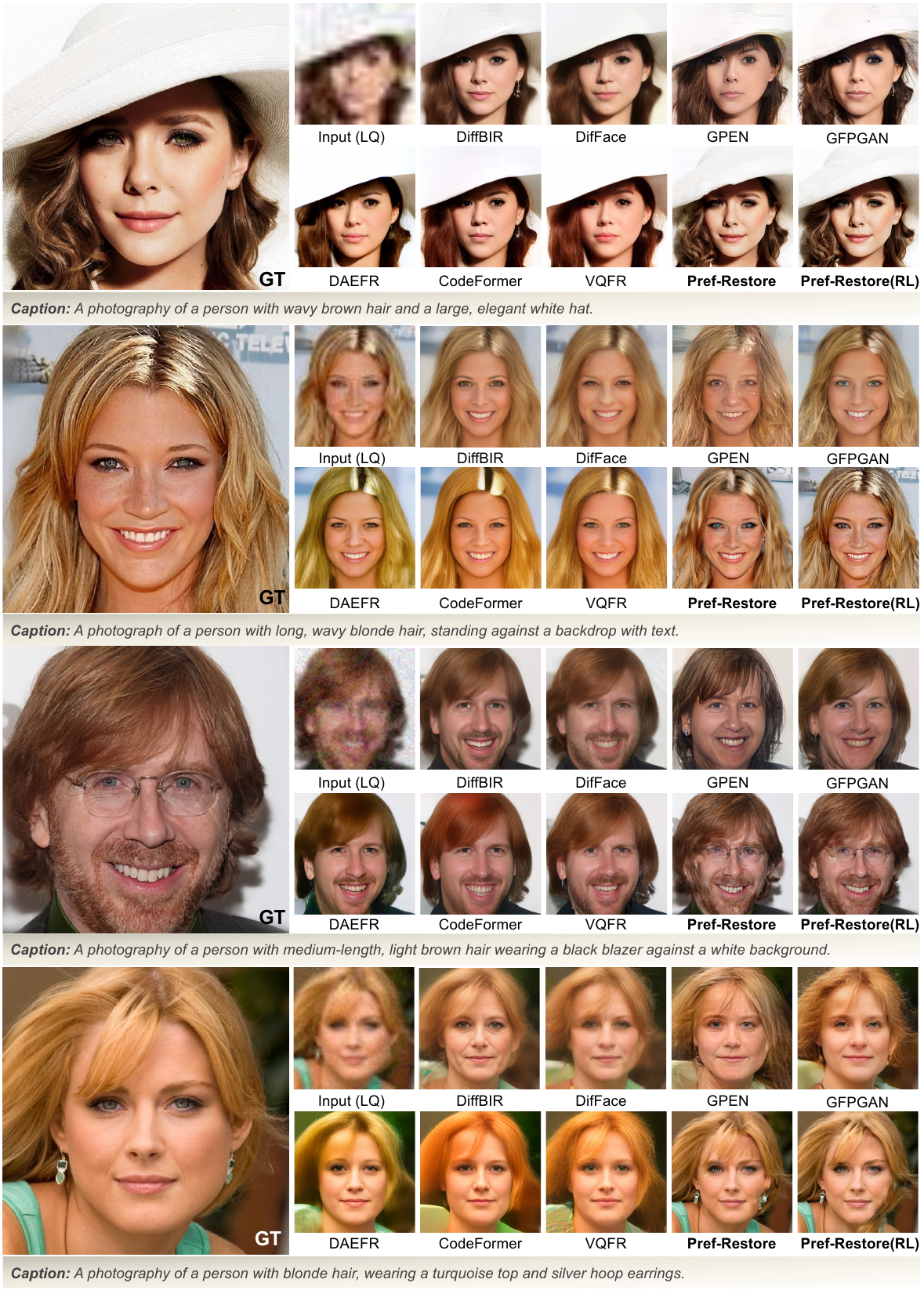}
    \vspace{-2mm}
    \caption{{\color{RevisionBlue}\textbf{Qualitative comparison of face restoration on the CelebA-HQ dataset.} We compare Pref-Restore and Pref-Restore (RL) with state-of-the-art methods. Pref-Restore better preserves identity-relevant structures, while Pref-Restore (RL) further improves local texture realism through fidelity-constrained preference optimization. Baselines such as CodeFormer and VQFR often produce over-smoothed details in complex regions.}}
    \label{fig:qualitative_celeba}
\end{figure*}

{\color{RevisionBlue}
We first conduct a qualitative comparison on the CelebA-HQ~\cite{karras2017celebA} dataset under controlled synthetic degradations. Fig.~\ref{fig:qualitative_celeba} compares Pref-Restore with several state-of-the-art methods, including GAN-based models (GFPGAN~\cite{wang2021gfpgan}, GPEN~\cite{yang2021gpen}) and codebook-based priors (DAEFR~\cite{tsai2023DAEFR}, CodeFormer~\cite{zhou2022codeformer}).
}

{\color{RevisionBlue}
\textbf{Stage-wise Qualitative Refinement.} 
Fig.~\ref{fig:qualitative_celeba} visualizes the complementary roles of Pref-Restore and Pref-Restore (RL). Pref-Restore, obtained after Stage~1 semantic--texture alignment, preserves identity-relevant structures such as eyeglass frames, eye gaze, and facial contours more faithfully than competing baselines. These details are often distorted or omitted by GAN-based methods such as GPEN or DR2, especially under severe degradation.

Building on this restoration base, Pref-Restore (RL) further refines perceptual texture quality through fidelity-constrained preference optimization. While maintaining the structural cues established in Stage~1, the RL-refined model produces cleaner facial textures, sharper contours, and more natural hair details. For example, in the row with blonde hair, CodeFormer tends to produce over-smoothed textures, whereas Pref-Restore (RL) recovers more realistic hair strands and local facial details without changing the overall identity structure.
}

{\color{RevisionBlue}
\textbf{Comparison with Baselines.} 
Traditional GAN-based methods (GFPGAN, GPEN) frequently suffer from identity drift or unnatural lighting artifacts under severe blur. Codebook-based methods such as CodeFormer and VQFR improve robustness, but often over-smooth high-frequency details, particularly in hair regions. In contrast, Pref-Restore benefits from Texture-to-Diffusion Alignment to preserve identity-oriented structure, and Pref-Restore (RL) further enhances local texture realism under fidelity constraints.
}

\subsection{The Role of Multi-modal Text Guidance}
\label{sec:text_guidance}

{\color{RevisionBlue}
We study multi-modal text guidance through three questions: whether the AR branch itself improves restoration, whether caption-conditioned training provides additional gains, and how useful text inferred from degraded inputs is when ground-truth captions are unavailable at inference. This design separates architectural benefit, training-time semantic supervision, and deployment-time text usability.

\textbf{Experimental Setup.} We retrain three Stage~1 models for 100k steps from the same initialization and with otherwise identical settings. \textit{Train-NoAR} removes the AR branch and text, serving as the baseline without semantic-token modeling. \textit{Train-Base} keeps the AR branch but uses no text, isolating the effect of the AR architecture itself. \textit{Train-Caption} keeps the AR branch and uses GT captions generated by Qwen2.5VL-32B~\cite{bai2025qwen2}, measuring the effect of caption-conditioned training. At inference, \textit{Test-NoText} removes runtime text, \textit{Test-GTCaption} provides an oracle upper bound with clean-image captions, \textit{Test-DegradedCaption} uses captions generated from the LQ input to approximate deployment, and \textit{Test-NoisyCaption} uses contradictory captions to test wrong-text sensitivity. We evaluate semantic and perceptual alignment with CLIPScore-I~\cite{radford2021clip}, CLIPScore-T~\cite{radford2021clip}, and DreamSim~\cite{fu2023dreamsim}, and restoration fidelity with LPIPS~\cite{zhang2018lpips}, ArcFace Deg~\cite{deng2019arcface}, and LMD~\cite{chen2018fsrnet}. Detailed prompts and caption construction are provided in the supplementary material.
}

\begin{table*}[!t]
    \centering
    \caption{{\color{RevisionBlue}\textbf{Three-model attribution and deployment robustness matrix for text guidance.} \textit{Train-NoAR} removes the AR module, \textit{Train-Base} keeps the AR module without text, and \textit{Train-Caption} uses text-conditioned training. At inference, \textit{Test-DegradedCaption} approximates deployment-time captioning from the LQ input, while \textit{Test-NoisyCaption} tests sensitivity to contradictory text.}}
    \vspace{-2mm}
    \label{tab:text_guidance}
    \resizebox{0.87\textwidth}{!}{
    \begin{tabular}{ll|ccc|ccc}
        \toprule
        \hfill \textbf{Training} & \textbf{Inference text} & \multicolumn{3}{c|}{\textbf{Semantic / Perceptual Alignment}} & \multicolumn{3}{c}{\textbf{Identity / Restoration Fidelity}} \\
        & & CLIPScore-I$\uparrow$ & CLIPScore-T$\uparrow$ & DreamSim$\uparrow$ & LPIPS$\downarrow$ & ArcFace Deg$\downarrow$ & LMD$\downarrow$ \\
        \midrule
        Train-NoAR & Test-NoText                & 0.8110 & 0.2869 & 0.8151 & 0.4492 & 63.0939  & 6.4610 \\
        \midrule
        Train-Base & Test-NoText                & 0.8162 & 0.2885 & 0.8236 & 0.4416 & 60.5767 & \textbf{6.1606} \\
        \midrule
        \multirow{4}{*}{Train-Caption} & Test-NoText    & 0.8163 & 0.2881 & 0.8229 & 0.4413 & 60.9876 & 6.2539 \\
        & Test-GTCaption          & \textbf{0.8372} & \textbf{0.2994} & \textbf{0.8532} & \textbf{0.4375} & \textbf{54.3705} & 6.4466 \\
        & Test-DegradedCaption    & 0.8134 & 0.2957 & 0.8342 & 0.4430 & 58.6068 & 7.1136 \\
        & Test-NoisyCaption       & 0.7784 & 0.2885 & 0.8004 & 0.4492 & 64.4612 & 7.9845 \\
        \bottomrule
    \end{tabular}
    }
    \vspace{-2mm}
\end{table*}

\begin{figure*}[!ht]
    \centering
    \includegraphics[width=0.95\textwidth]{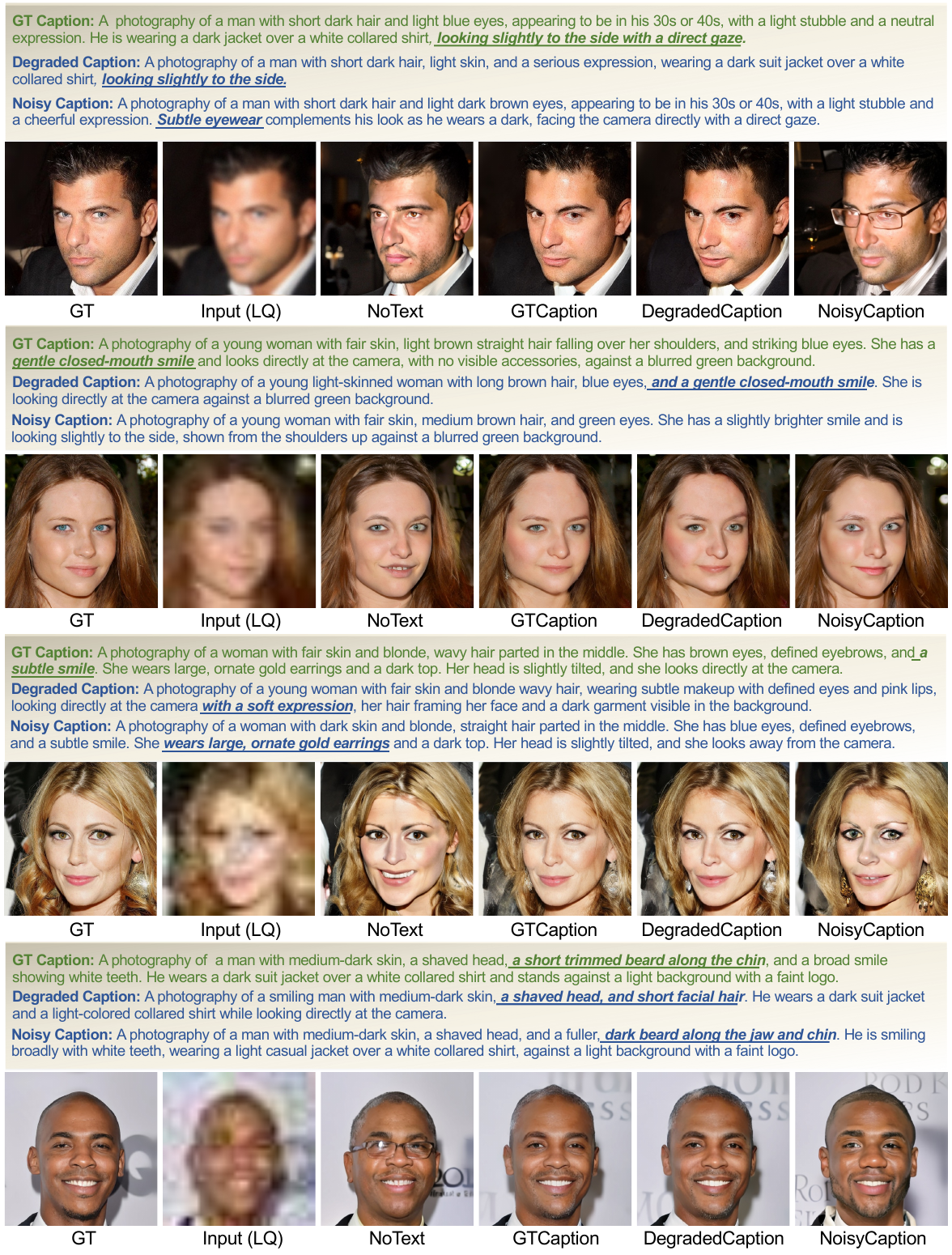}
    \vspace{-2mm}
    \caption{\textbf{Qualitative attribution for the text pathway.} We include the caption-trained model under different inference-time text conditions. The examples mainly show that accurate text provides useful identity guidance, degraded-input captions can still help, and contradictory noisy captions may steer the restoration away from the target identity.}
    \label{fig:r4_text_ablation}
\end{figure*}

{\color{RevisionBlue}
\textbf{Quantitative Analysis.} Table~\ref{tab:text_guidance} answers the three questions above. (1) AR branch. \textit{Train-Base} improves over \textit{Train-NoAR} on all metrics, showing that the AR branch itself provides a useful semantic pathway even without text and that the gain is not solely a caption effect. (2) Caption-conditioned training. \textit{Train-Caption / Test-GTCaption} achieves the best semantic scores and the strongest ArcFace Deg, confirming that accurate captions are actively used as identity-relevant guidance rather than as a passive side input. (3) Deployment-time text. Compared with \textit{Test-NoText}, \textit{Test-DegradedCaption} improves CLIPScore-T, DreamSim, and ArcFace Deg with only a small LPIPS change, suggesting that captions inferred from degraded inputs can still provide actionable semantic cues when GT captions are unavailable. Meanwhile, \textit{Test-NoisyCaption} degrades semantic and fidelity metrics, indicating that runtime text is useful but quality-dependent.

\textbf{Qualitative Analysis.} Fig.~\ref{fig:r4_text_ablation} provides representative examples for the caption-trained model under different inference-time text conditions. Accurate captions provide useful identity guidance, while captions inferred from degraded inputs can still help when GT captions are unavailable. In contrast, contradictory noisy captions may steer the restoration away from the target identity, consistent with the degradation of semantic and fidelity metrics in Table~\ref{tab:text_guidance}. We therefore treat text as a useful but quality-dependent semantic condition rather than an idealized assumption.
}

\subsection{\texorpdfstring{\textcolor{RevisionBlue}{Fidelity-Constrained Preference Optimization}}{Fidelity-Constrained Preference Optimization}}
\label{sec:diffusionnft_results}


\begin{table*}[!ht]
    \caption{{\color{RevisionBlue}\textbf{Reward-composition ablation for Stage~2 on CelebA-Test.} Starting from the same Stage~1 checkpoint, we progressively add restoration-specific reward terms to the aesthetic reward. Values in parentheses denote absolute changes relative to \textit{No Stage~2}; \red{red} indicates a favorable change for the corresponding metric direction, while \blue{blue} indicates deterioration.}}
    \label{tab:stage2_reward_ablation}
    \centering
    \resizebox{\textwidth}{!}{
    \begin{tabular}{l|ccc|ccc}
        \toprule
        \hfill \textbf{Stage~2 reward} & \multicolumn{3}{c|}{\textbf{Perceptual / General Quality}} & \multicolumn{3}{c}{\textbf{Identity / Geometry Fidelity}} \\
        \textbf{Composition} & MUSIQ$\uparrow$ & CLIPIQA+$\uparrow$ & MANIQA$\uparrow$ & LPIPS$\downarrow$ & ArcFace Deg$\downarrow$ & LMD$\downarrow$ \\
        \midrule
        No Stage~2 (= Stage~1.1 + 1.2) & 74.1902 & 0.6526 & 0.4895 & 0.4175 & 54.0623 & 5.1337 \\
        \midrule
        Aesthetic-only ($R_{\mathrm{aes}}$) & 76.3296 \red{(+2.139)} & 0.6997 \red{(+0.047)} & 0.5810 \red{(+0.092)} & 0.5099 \blue{(+0.092)} & 73.9882 \blue{(+19.926)} & 8.3588 \blue{(+3.225)} \\
        + Identity ($R_{\mathrm{aes}}, R_{\mathrm{id}}$) & 77.3114 \red{(+3.121)} & 0.7314 \red{(+0.079)} & 0.5819 \red{(+0.092)} & 0.3882 \red{(-0.029)} & 47.1055 \red{(-6.957)} & 3.4407 \red{(-1.693)} \\
        + Identity + LPIPS ($R_{\mathrm{aes}}, R_{\mathrm{id}}, R_{\mathrm{lpips}}$) & 76.6035 \red{(+2.413)} & 0.7179 \red{(+0.065)} & 0.5589 \red{(+0.069)} & 0.3788 \red{(-0.039)} & 47.0899 \red{(-6.972)} & 3.4114 \red{(-1.722)} \\
        \rowcolor{lightblue}
        + Identity + LPIPS + Geometry ($R_{\mathrm{aes}}, R_{\mathrm{id}}, R_{\mathrm{lpips}}, R_{\mathrm{geo}}$) & 75.5061 \red{(+1.316)} & 0.6989 \red{(+0.046)} & 0.5216 \red{(+0.032)} & 0.3966 \red{(-0.021)} & 49.6654 \red{(-4.397)} & 4.0559 \red{(-1.078)} \\
        \bottomrule
    \end{tabular}
    }
    \vspace{-2mm}
\end{table*}

\begin{figure*}[!t]
    \centering
    \includegraphics[width=\textwidth]{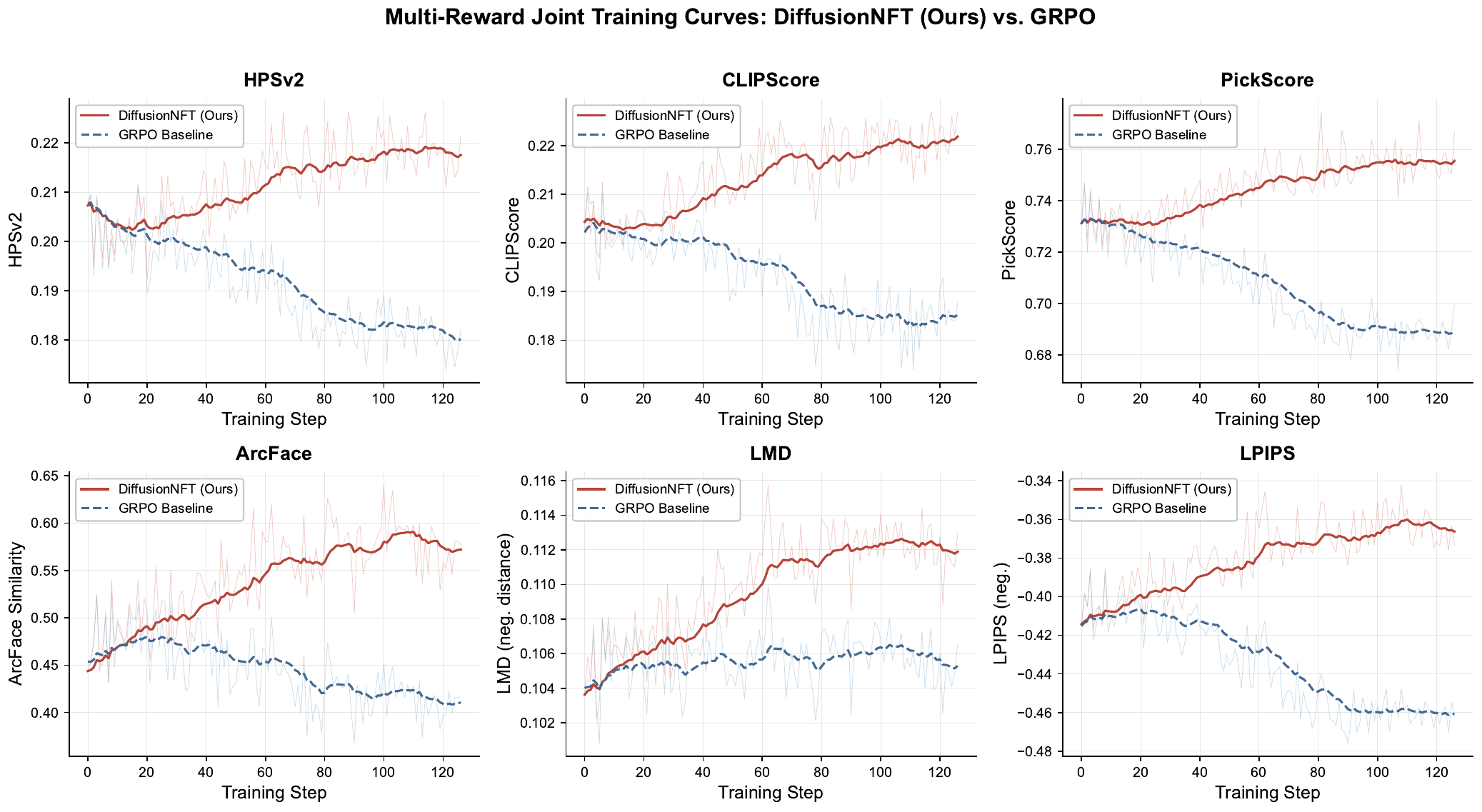}
    \vspace{-2mm}
    \caption{{\color{RevisionBlue}\textbf{Multi-reward training curves for Stage~2 optimizer comparison.} DiffusionNFT and GRPO start from the same Stage~1 checkpoint and optimize the same six reward signals: HPSv2, CLIPScore, PickScore, ArcFace, LMD, and LPIPS. DiffusionNFT maintains stable improvements across quality and fidelity rewards, while GRPO becomes unstable under this joint objective.}}
    \label{fig:stage2_grpo_curves}
    \vspace{-2mm}
\end{figure*}

{\color{RevisionBlue}
Stage~2 evaluates whether reward-guided fine-tuning can improve BFR without drifting away from the target identity. We therefore instantiate the forward-process optimizer of DiffusionNFT~\cite{zheng2025diffusionnft} with the fidelity-constrained reward in Eq.~\eqref{eq:pref_reward}. This section studies two aspects: how the added identity, LPIPS, and geometry rewards reshape the quality--fidelity trade-off, and whether the resulting multi-objective reward can be optimized more stably than a GRPO-style reverse-process baseline~\cite{shao2024grpo,liu2025flowgrpo}.

\textbf{Reward Composition.} Table~\ref{tab:stage2_reward_ablation} shows that reward design is decisive for Stage~2 under a fixed Stage~1 checkpoint. The aesthetic-only reward improves no-reference quality metrics, but it also worsens LPIPS, ArcFace Deg, and LMD, indicating identity and geometry drift. Adding the identity reward reverses this trend and improves both perceptual quality and restoration fidelity over the no-Stage~2 baseline. LPIPS and geometry terms further constrain the output toward reference similarity and facial-structure consistency. In this ablation, the full reward provides a conservative BFR operating point: it improves all reported metrics over no Stage~2, while avoiding the fidelity degradation caused by aesthetic-only preference transfer. A denser sweep of the fidelity weight is provided in the supplementary material.
}

{\color{RevisionBlue}
\textbf{Optimization Dynamics.} A richer reward is useful only if it can be optimized stably. We therefore compare DiffusionNFT with a GRPO-style baseline under the same Stage~1 checkpoint and the same six reward signals. As shown in Fig.~\ref{fig:stage2_grpo_curves}, DiffusionNFT steadily improves the aesthetic and fidelity rewards, whereas the GRPO baseline becomes unstable under the joint objective. This comparison supports the use of forward-process preference optimization for BFR: the reward specifies the restoration target, and the optimizer must preserve stable progress when perceptual and fidelity objectives compete. Additional convergence curves and downstream GRPO results are reported in the supplementary material.
}
\subsection{\texorpdfstring{\textcolor{RevisionBlue}{Ablation Study: Staged Semantic--Texture Alignment}}{Ablation Study: Staged Semantic--Texture Alignment}}
\label{sec:ablation_step2}

{\color{RevisionBlue}
Stage~1 contains two coupled but distinct objectives: \textit{Semantic-to-Diffusion Alignment} first stabilizes the AR semantic tokens as diffusion conditions, while \textit{Texture-to-Diffusion Alignment} then adapts the diffusion branch to identity-relevant texture cues. To test both the contribution of each sub-stage and the need for sequential optimization, we compare three controlled settings in Table~\ref{tab:ablation_step2}. \textit{Stage~1.1 only} is trained for 50k steps. \textit{Stage~1.1 + Stage~1.2} is trained sequentially for 50k+50k steps. \textit{Joint optimization} is trained end-to-end for 100k steps, matching the total budget of the staged setting.

\textbf{Quantitative Analysis.} Stage~1.1 already produces plausible restorations, showing that semantic alignment provides a useful generative prior. However, it remains weak on identity-sensitive fidelity, with LPIPS $0.5272$, LMD $8.6819$, and ArcFace Deg $74.5286$. Adding Stage~1.2 substantially improves these restoration metrics, reducing LPIPS to $0.4413$, LMD to $6.2539$, and ArcFace Deg to $60.9876$. This confirms that texture-level alignment is the key step that turns semantic plausibility into identity-faithful restoration.

The comparison with joint optimization further shows why the two steps should be separated. Under the same 100k-step budget, staged training outperforms joint optimization on LPIPS, FID(HQ), LMD, ArcFace Deg, and all three face no-reference metrics. This suggests that early diffusion updates benefit from a stabilized AR semantic condition; when semantic and texture alignment are optimized jointly from the beginning, the diffusion branch receives less stable conditioning signals and can prematurely adapt to noisy semantic anchors, eventually converging to a weaker fidelity operating point. Thus, the hierarchy is not only beneficial component-wise, but also structurally important for training.
}

\begin{table*}[!ht]
    \centering
    \caption{{\color{RevisionBlue}\textbf{Ablation study of staged semantic--texture alignment on CelebA-Test.} We compare Stage~1.1 only, sequential Stage~1.1+Stage~1.2, and joint optimization under a matched training budget. Joint optimization is trained end-to-end for 100k steps; the staged model uses 50k steps for Stage~1.1 and 50k for Stage~1.2.}}
    \label{tab:ablation_step2}
    \resizebox{\textwidth}{!}{
    \begin{tabular}{l|cccccc|ccccc}
        \toprule
        \hfill \textbf{CelebA-Test} & \multicolumn{6}{c|}{\textbf{General Image Quality Evaluation}} & \multicolumn{5}{c}{\textbf{Face Image Quality Evaluation}} \\
        \textbf{Configuration} & LPIPS$\downarrow$ & FID(HQ)$\downarrow$ & FID(FFHQ)$\downarrow$ & MUSIQ$\uparrow$ & CLIPIQA+$\uparrow$ & MANIQA$\uparrow$ & LMD$\downarrow$ & Deg$\downarrow$ & topiq$\uparrow$ & topiq-swin$\uparrow$ & DSL-FIQA$\uparrow$ \\
        \midrule
        Stage~1.1 only          & 0.5272 & 23.2199 & 45.5583 & \textbf{76.3296} & \textbf{0.6968} & \textbf{0.5353} & 8.6819 & 74.5286 & \textbf{0.7983} & \textbf{0.8670} & 0.8573 \\
        \rowcolor{lightblue}
        \textbf{Stage~1.1 + Stage~1.2}   & \textbf{0.4413} & \textbf{15.9407} & 44.9123 & 75.6157 & 0.6714 & 0.5233 & \textbf{6.2539} & \textbf{60.9876} & 0.7964 & 0.8654 & \textbf{0.8591} \\
        Joint optimization      & 0.4457 & 17.8083 & \textbf{40.1371} & 74.8526 & 0.6523 & 0.5040 & 6.3630 & 62.1582 & 0.7679 & 0.8504 & 0.8369 \\
        \bottomrule
    \end{tabular}
    }
\end{table*}

\begin{table*}[!ht]
    \centering
    \caption{{\color{RevisionBlue}\textbf{Seed-consistency comparison under $N=16$ restoration runs.} For each test image, we compute the across-seed standard deviation. Lower values indicate lower restoration uncertainty.}}
    \label{tab:seed_consistency}
    \resizebox{0.75\linewidth}{!}{
    \begin{tabular}{l|ccc|cc}
        \toprule
        \hfill \textbf{Methods} & \multicolumn{3}{c|}{\textbf{Preference Stability}} & \multicolumn{2}{c}{\textbf{Identity / Geometry Stability}} \\
        & std(HPSv2)$\downarrow$ & std(PickScore)$\downarrow$ & std(CLIPScore)$\downarrow$ & std(LMD)$\downarrow$ & std(ArcFace Deg)$\downarrow$\\
        \midrule
        Pref-Restore (Stage~1) & 0.0122 & 0.0182 & 0.0185 & 1.2401 & 3.2516 \\
        \rowcolor{lightblue}
        \textbf{Pref-Restore (RL)} & \textbf{0.0051} & \textbf{0.0083} & \textbf{0.0090} & \textbf{0.3592} & \textbf{1.7214} \\
        \bottomrule
    \end{tabular}
    }
\end{table*}

\begin{figure*}[!t]
    \centering
    \includegraphics[width=\linewidth]{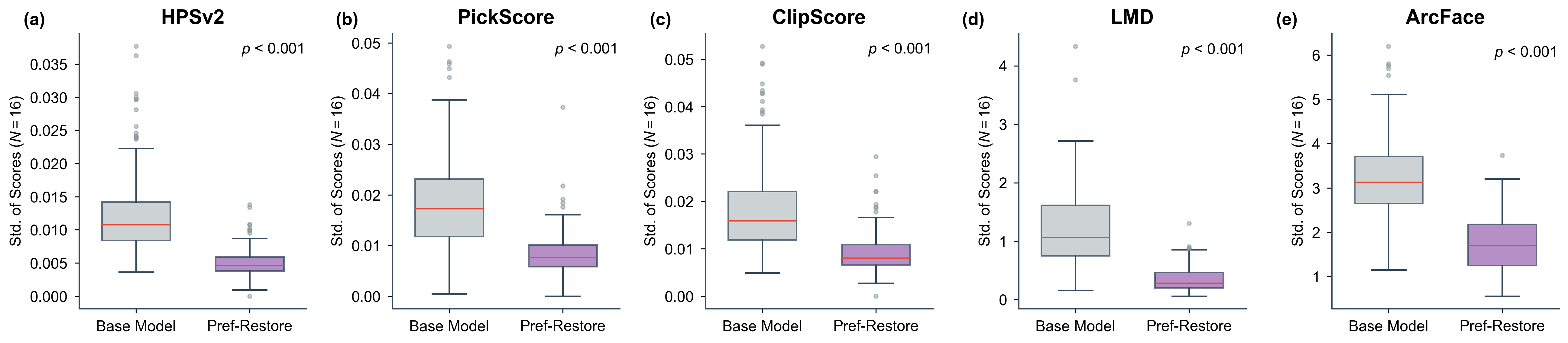}
    \caption{{\color{RevisionBlue}\textbf{Per-image seed-variance distributions.} For each image, we compute the across-seed standard deviation over $N=16$ restoration runs for HPSv2, PickScore, CLIPScore, LMD, and ArcFace Deg. The base model denotes Pref-Restore after Stage~1, while Pref-Restore (RL) denotes the final Stage~2 model. Pref-Restore (RL) yields tighter distributions across preference, identity, and geometry metrics.}}
    \label{fig:stochasticity_boxplot}
\end{figure*}

\subsection{\texorpdfstring{\textcolor{RevisionBlue}{Deterministic Analysis: Reduced Restoration Uncertainty}}{Deterministic Analysis: Reduced Restoration Uncertainty}}
\label{sec:determinism_analysis}



{\color{RevisionBlue}
Blind face restoration is intrinsically uncertain: the same degraded input may admit multiple plausible faces when high-frequency identity cues are missing or heavily corrupted. We therefore evaluate whether Pref-Restore reduces \textit{seed-to-seed restoration uncertainty} under a fixed degraded input. For each test image, we run $N=16$ independent restorations with different random seeds and compute the across-seed standard deviation. This analysis covers preference stability (HPSv2, PickScore, and CLIPScore), facial geometry stability (LMD), and identity stability (ArcFace Deg), giving a more complete view of restoration consistency.

\textbf{Quantitative Seed Consistency.} Table~\ref{tab:seed_consistency} shows that Stage~2 fidelity-constrained preference optimization reduces variance on both preference and face-relevant metrics. Compared with the Stage~1 model, Pref-Restore (RL) lowers std(HPSv2) from $0.0122$ to $0.0051$, std(PickScore) from $0.0182$ to $0.0083$, and std(CLIPScore) from $0.0185$ to $0.0090$. More importantly for BFR, the same trend holds for identity and geometry: std(LMD) decreases from $1.2401$ to $0.3592$, and std(ArcFace Deg) decreases from $3.2516$ to $1.7214$.

\textbf{Distributional Evidence.} Fig.~\ref{fig:stochasticity_boxplot} further visualizes per-image standard-deviation distributions across the test set. The RL-refined model consistently shifts the distributions toward lower variance across all five metrics, indicating more concentrated restoration behavior under repeated sampling. Together with Table~\ref{tab:seed_consistency}, this shows that Pref-Restore (RL) is less sensitive to sampling randomness and more consistent in identity and facial structure.
}

\section{Conclusion}
\label{sec:conclusion}

{\color{RevisionBlue}
In this paper, we present Pref-Restore, a hierarchical blind face restoration framework that combines staged semantic--texture alignment with fidelity-constrained preference optimization. Stage~1 first aligns AR semantic tokens with the diffusion conditioning space and then injects texture-level cues for identity-preserving restoration. Stage~2 further refines the restoration trajectory with a face-aware reward that balances perceptual quality, identity similarity, reference similarity, and facial geometry. Extensive experiments show that Pref-Restore achieves competitive restoration quality on synthetic and real-world benchmarks, while the ablation studies validate the necessity of staged training, realistic text guidance, and fidelity-aware preference optimization. Seed-consistency analysis further indicates reduced restoration uncertainty across preference, identity, and geometry metrics. 
}

\section*{Acknowledgments}
This work was supported in part by National Natural Science Foundation of China (62501020, 82371112); in part by the Capital's Funds for HealthImprovement and Research (2026-1-2151); in part by the Science Foundation of Peking University Cancer Hospital (JC202505); in part by the China National Postdoctoral Program for Innovative Talents (BX20250368).

\nolinenumbers
\bibliographystyle{IEEEtran}
\bibliography{main}

@String(ECCV= {Eur. Conf. Comput. Vis.})

@String(AAAI = {AAAI})

@String(ECCV  = {ECCV})

@inproceedings{ledig2017SRGAN,
  title={Photo-realistic single image super-resolution using a generative adversarial network},
  author={Ledig, Christian and Theis, Lucas and Husz{\'a}r, Ferenc and Caballero, Jose and Cunningham, Andrew and Acosta, Alejandro and Aitken, Andrew and Tejani, Alykhan and Totz, Johannes and Wang, Zehan and others},
  booktitle={Proceedings of the IEEE conference on computer vision and pattern recognition},
  pages={4681--4690},
  year={2017}
}

@article{goodfellow2014gan,
  title={Generative adversarial nets},
  author={Goodfellow, Ian and Pouget-Abadie, Jean and Mirza, Mehdi and Xu, Bing and Warde-Farley, David and Ozair, Sherjil and Courville, Aaron and Bengio, Yoshua},
  journal={Advances in neural information processing systems},
  volume={27},
  year={2014}
}

@article{ho2020ddpm,
  title={Denoising diffusion probabilistic models},
  author={Ho, Jonathan and Jain, Ajay and Abbeel, Pieter},
  journal={Advances in neural information processing systems},
  volume={33},
  pages={6840--6851},
  year={2020}
}

@inproceedings{gu2022vqfr,
  title={Vqfr: Blind face restoration with vector-quantized dictionary and parallel decoder},
  author={Gu, Yuchao and Wang, Xintao and Xie, Liangbin and Dong, Chao and Li, Gen and Shan, Ying and Cheng, Ming-Ming},
  booktitle={European Conference on Computer Vision},
  pages={126--143},
  year={2022},
  organization={Springer}
}

@article{zhou2022codeformer,
  title={Towards robust blind face restoration with codebook lookup transformer},
  author={Zhou, Shangchen and Chan, Kelvin and Li, Chongyi and Loy, Chen Change},
  journal={Advances in Neural Information Processing Systems},
  volume={35},
  pages={30599--30611},
  year={2022}
}

@article{tsai2023DAEFR,
  title={Dual associated encoder for face restoration},
  author={Tsai, Yu-Ju and Liu, Yu-Lun and Qi, Lu and Chan, Kelvin CK and Yang, Ming-Hsuan},
  journal={arXiv preprint arXiv:2308.07314},
  year={2023}
}

@inproceedings{chen2021progressive,
  title={Progressive semantic-aware style transformation for blind face restoration},
  author={Chen, Chaofeng and Li, Xiaoming and Yang, Lingbo and Lin, Xianhui and Zhang, Lei and Wong, Kwan-Yee K},
  booktitle={Proceedings of the IEEE/CVF conference on computer vision and pattern recognition},
  pages={11896--11905},
  year={2021}
}

@inproceedings{chen2018fsrnet,
  title={Fsrnet: End-to-end learning face super-resolution with facial priors},
  author={Chen, Yu and Tai, Ying and Liu, Xiaoming and Shen, Chunhua and Yang, Jian},
  booktitle={Proceedings of the IEEE conference on computer vision and pattern recognition},
  pages={2492--2501},
  year={2018}
}

@inproceedings{hu2020face,
  title={Face super-resolution guided by 3d facial priors},
  author={Hu, Xiaobin and Ren, Wenqi and LaMaster, John and Cao, Xiaochun and Li, Xiaoming and Li, Zechao and Menze, Bjoern and Liu, Wei},
  booktitle={Computer Vision--ECCV 2020: 16th European Conference, Glasgow, UK, August 23--28, 2020, Proceedings, Part IV 16},
  pages={763--780},
  year={2020},
  organization={Springer}
}

@inproceedings{dogan2019exemplar,
  title={Exemplar guided face image super-resolution without facial landmarks},
  author={Dogan, Berk and Gu, Shuhang and Timofte, Radu},
  booktitle={Proceedings of the IEEE/CVF conference on computer vision and pattern recognition workshops},
  pages={0--0},
  year={2019}
}

@inproceedings{li2020blind,
  title={Blind face restoration via deep multi-scale component dictionaries},
  author={Li, Xiaoming and Chen, Chaofeng and Zhou, Shangchen and Lin, Xianhui and Zuo, Wangmeng and Zhang, Lei},
  booktitle={European conference on computer vision},
  pages={399--415},
  year={2020},
  organization={Springer}
}

@inproceedings{karras2019stylegan,
  title={A style-based generator architecture for generative adversarial networks},
  author={Karras, Tero and Laine, Samuli and Aila, Timo},
  booktitle={Proceedings of the IEEE/CVF conference on computer vision and pattern recognition},
  pages={4401--4410},
  year={2019}
}

@inproceedings{gu2020image,
  title={Image processing using multi-code gan prior},
  author={Gu, Jinjin and Shen, Yujun and Zhou, Bolei},
  booktitle={Proceedings of the IEEE/CVF conference on computer vision and pattern recognition},
  pages={3012--3021},
  year={2020}
}

@inproceedings{menon2020pulse,
  title={Pulse: Self-supervised photo upsampling via latent space exploration of generative models},
  author={Menon, Sachit and Damian, Alexandru and Hu, Shijia and Ravi, Nikhil and Rudin, Cynthia},
  booktitle={Proceedings of the ieee/cvf conference on computer vision and pattern recognition},
  pages={2437--2445},
  year={2020}
}

@inproceedings{yang2021gpen,
  title={Gan prior embedded network for blind face restoration in the wild},
  author={Yang, Tao and Ren, Peiran and Xie, Xuansong and Zhang, Lei},
  booktitle={Proceedings of the IEEE/CVF conference on computer vision and pattern recognition},
  pages={672--681},
  year={2021}
}

@inproceedings{wang2021gfpgan,
  title={Towards real-world blind face restoration with generative facial prior},
  author={Wang, Xintao and Li, Yu and Zhang, Honglun and Shan, Ying},
  booktitle={Proceedings of the IEEE/CVF conference on computer vision and pattern recognition},
  pages={9168--9178},
  year={2021}
}

@inproceedings{wang2023dr2,
  title={Dr2: Diffusion-based robust degradation remover for blind face restoration},
  author={Wang, Zhixin and Zhang, Ziying and Zhang, Xiaoyun and Zheng, Huangjie and Zhou, Mingyuan and Zhang, Ya and Wang, Yanfeng},
  booktitle={Proceedings of the IEEE/CVF Conference on Computer Vision and Pattern Recognition},
  pages={1704--1713},
  year={2023}
}

@inproceedings{lin2023diffbir,
  title={{DiffBIR}: Toward Blind Image Restoration with Generative Diffusion Prior},
  author={Lin, Xinqi and He, Jingwen and Chen, Ziyan and Lyu, Zhaoyang and Dai, Bo and Yu, Fanghua and Qiao, Yu and Ouyang, Wanli and Dong, Chao},
  booktitle={European Conference on Computer Vision},
  pages={430--448},
  year={2024}
}

@article{karras2017celebA,
  title={Progressive Growing of GANs for Improved Quality, Stability, and Variation},
  author={Karras, Tero},
  journal={arXiv preprint arXiv:1710.10196},
  year={2017}
}

@inproceedings{huang2008lfw,
  title={Labeled faces in the wild: A database forstudying face recognition in unconstrained environments},
  author={Huang, Gary B and Mattar, Marwan and Berg, Tamara and Learned-Miller, Eric},
  booktitle={Workshop on faces in'Real-Life'Images: detection, alignment, and recognition},
  year={2008}
}

@inproceedings{yang2016wider,
  title={Wider face: A face detection benchmark},
  author={Yang, Shuo and Luo, Ping and Loy, Chen-Change and Tang, Xiaoou},
  booktitle={Proceedings of the IEEE conference on computer vision and pattern recognition},
  pages={5525--5533},
  year={2016}
}

@inproceedings{zhang2018lpips,
  title={The unreasonable effectiveness of deep features as a perceptual metric},
  author={Zhang, Richard and Isola, Phillip and Efros, Alexei A and Shechtman, Eli and Wang, Oliver},
  booktitle={Proceedings of the IEEE conference on computer vision and pattern recognition},
  pages={586--595},
  year={2018}
}

@article{heusel2017fid,
  title={Gans trained by a two time-scale update rule converge to a local nash equilibrium},
  author={Heusel, Martin and Ramsauer, Hubert and Unterthiner, Thomas and Nessler, Bernhard and Hochreiter, Sepp},
  journal={Advances in neural information processing systems},
  volume={30},
  year={2017}
}

@inproceedings{deng2019arcface,
  title={Arcface: Additive angular margin loss for deep face recognition},
  author={Deng, Jiankang and Guo, Jia and Xue, Niannan and Zafeiriou, Stefanos},
  booktitle={Proceedings of the IEEE/CVF conference on computer vision and pattern recognition},
  pages={4690--4699},
  year={2019}
}

@inproceedings{yang2022maniqa,
  title={Maniqa: Multi-dimension attention network for no-reference image quality assessment},
  author={Yang, Sidi and Wu, Tianhe and Shi, Shuwei and Lao, Shanshan and Gong, Yuan and Cao, Mingdeng and Wang, Jiahao and Yang, Yujiu},
  booktitle={Proceedings of the IEEE/CVF Conference on Computer Vision and Pattern Recognition},
  pages={1191--1200},
  year={2022}
}

@inproceedings{ke2021musiq,
  title={Musiq: Multi-scale image quality transformer},
  author={Ke, Junjie and Wang, Qifei and Wang, Yilin and Milanfar, Peyman and Yang, Feng},
  booktitle={Proceedings of the IEEE/CVF international conference on computer vision},
  pages={5148--5157},
  year={2021}
}

@inproceedings{wang2023clipiqa,
  title={Exploring clip for assessing the look and feel of images},
  author={Wang, Jianyi and Chan, Kelvin CK and Loy, Chen Change},
  booktitle={Proceedings of the AAAI Conference on Artificial Intelligence},
  volume={37},
  number={2},
  pages={2555--2563},
  year={2023}
}

@article{potlapalli2023promptir,
  title={Promptir: Prompting for all-in-one image restoration},
  author={Potlapalli, Vaishnav and Zamir, Syed Waqas and Khan, Salman H and Shahbaz Khan, Fahad},
  journal={Advances in Neural Information Processing Systems},
  volume={36},
  pages={71275--71293},
  year={2023}
}

@inproceedings{conde2024instructir,
  title={Instructir: High-quality image restoration following human instructions},
  author={Conde, Marcos V and Geigle, Gregor and Timofte, Radu},
  booktitle={European Conference on Computer Vision},
  pages={1--21},
  year={2024},
  organization={Springer}
}

@article{luo2023controlling,
  title={Controlling vision-language models for multi-task image restoration},
  author={Luo, Ziwei and Gustafsson, Fredrik K and Zhao, Zheng and Sj{\"o}lund, Jens and Sch{\"o}n, Thomas B},
  journal={arXiv preprint arXiv:2310.01018},
  year={2023}
}

@inproceedings{yu2018crafting,
 author = {Yu, Ke and Dong, Chao and Lin, Liang and Loy, Chen Change},
 title = {Crafting a Toolchain for Image Restoration by Deep Reinforcement Learning},
 booktitle = {Proceedings of IEEE Conference on Computer Vision and Pattern Recognition},
 pages={2443--2452},
 year = {2018} 
}

@article{yu2021path,
title={Path-Restore: Learning Network Path Selection for Image Restoration},
author={Yu, Ke and Wang, Xintao and Dong, Chao and Tang, Xiaoou and Loy, Chen Change},
journal={IEEE Transactions on Pattern Analysis and Machine Intelligence},
year={2021},
publisher={IEEE}
}

@article{wu2025diffusionreward,
  title={DiffusionReward: Enhancing Blind Face Restoration through Reward Feedback Learning},
  author={Wu, Bin and Wang, Wei and Liu, Yahui and Li, Zixiang and Zhao, Yao},
  journal={arXiv preprint arXiv:2505.17910},
  year={2025}
}

@article{cai2025dspo,
  title={DSPO: Direct Semantic Preference Optimization for Real-World Image Super-Resolution},
  author={Cai, Miaomiao and Li, Simiao and Li, Wei and Huang, Xudong and Chen, Hanting and Hu, Jie and Wang, Yunhe},
  journal={arXiv preprint arXiv:2504.15176},
  year={2025}
}

@article{liu2025irpo,
  title={IRPO: Boosting Image Restoration via Post-training GRPO},
  author={Liu, Haoxuan Xu Yi and Jiang, Boyuan and Peng, Jinlong and Luo, Donghao and Hu, Xiaobin and Yan, Shuicheng and Li, Haoang},
  journal={arXiv preprint arXiv:2512.00814},
  year={2025}
}

@article{li2025test,
  title={Test-Time Preference Optimization for Image Restoration},
  author={Li, Bingchen and Li, Xin and Xu, Jiaqi and Guo, Jiaming and Li, Wenbo and Pei, Renjing and Chen, Zhibo},
  journal={arXiv preprint arXiv:2511.19169},
  year={2025}
}

@article{zheng2025diffusionnft,
  title={Diffusionnft: Online diffusion reinforcement with forward process},
  author={Zheng, Kaiwen and Chen, Huayu and Ye, Haotian and Wang, Haoxiang and Zhang, Qinsheng and Jiang, Kai and Su, Hang and Ermon, Stefano and Zhu, Jun and Liu, Ming-Yu},
  journal={arXiv preprint arXiv:2509.16117},
  year={2025}
}

@article{chen2025blip3o,
  title={Blip3o-next: Next frontier of native image generation},
  author={Chen, Jiuhai and Xue, Le and Xu, Zhiyang and Pan, Xichen and Yang, Shusheng and Qin, Can and Yan, An and Zhou, Honglu and Chen, Zeyuan and Huang, Lifu and others},
  journal={arXiv preprint arXiv:2510.15857},
  year={2025}
}

@article{tschannen2025siglip,
  title={Siglip 2: Multilingual vision-language encoders with improved semantic understanding, localization, and dense features},
  author={Tschannen, Michael and Gritsenko, Alexey and Wang, Xiao and Naeem, Muhammad Ferjad and Alabdulmohsin, Ibrahim and Parthasarathy, Nikhil and Evans, Talfan and Beyer, Lucas and Xia, Ye and Mustafa, Basil and others},
  journal={arXiv preprint arXiv:2502.14786},
  year={2025}
}

@article{xie2025sana,
  title={Sana 1.5: Efficient scaling of training-time and inference-time compute in linear diffusion transformer},
  author={Xie, Enze and Chen, Junsong and Zhao, Yuyang and Yu, Jincheng and Zhu, Ligeng and Wu, Chengyue and Lin, Yujun and Zhang, Zhekai and Li, Muyang and Chen, Junyu and others},
  journal={arXiv preprint arXiv:2501.18427},
  year={2025}
}

@inproceedings{karras2019style,
  title={A style-based generator architecture for generative adversarial networks},
  author={Karras, Tero and Laine, Samuli and Aila, Timo},
  booktitle={Proceedings of the IEEE/CVF conference on computer vision and pattern recognition},
  pages={4401--4410},
  year={2019}
}

@misc{pyiqa,
  title={{IQA-PyTorch}: PyTorch Toolbox for Image Quality Assessment},
  author={Chaofeng Chen and Jiadi Mo},
  year={2022},
  howpublished = "[Online]. Available: \url{https://github.com/chaofengc/IQA-PyTorch}"
}

@article{wang2023restoreformer++,
  title={Restoreformer++: Towards real-world blind face restoration from undegraded key-value pairs},
  author={Wang, Zhouxia and Zhang, Jiawei and Chen, Tianshui and Wang, Wenping and Luo, Ping},
  journal={IEEE Transactions on Pattern Analysis and Machine Intelligence},
  volume={45},
  number={12},
  pages={15462--15476},
  year={2023},
  publisher={IEEE}
}

@article{yue2024difface,
  title={Difface: Blind face restoration with diffused error contraction},
  author={Yue, Zongsheng and Loy, Chen Change},
  journal={IEEE Transactions on Pattern Analysis and Machine Intelligence},
  year={2024},
  publisher={IEEE}
}

@article{bai2025qwen2,
  title={Qwen2. 5-vl technical report},
  author={Bai, Shuai and Chen, Keqin and Liu, Xuejing and Wang, Jialin and Ge, Wenbin and Song, Sibo and Dang, Kai and Wang, Peng and Wang, Shijie and Tang, Jun and others},
  journal={arXiv preprint arXiv:2502.13923},
  year={2025}
}

@inproceedings{radford2021clip,
  title={Learning transferable visual models from natural language supervision},
  author={Radford, Alec and Kim, Jong Wook and Hallacy, Chris and Ramesh, Aditya and Goh, Gabriel and Agarwal, Sandhini and Sastry, Girish and Askell, Amanda and Mishkin, Pamela and Clark, Jack and others},
  booktitle={International conference on machine learning},
  pages={8748--8763},
  year={2021},
  organization={PmLR}
}

@article{fu2023dreamsim,
  title={Dreamsim: Learning new dimensions of human visual similarity using synthetic data},
  author={Fu, Stephanie and Tamir, Netanel and Sundaram, Shobhita and Chai, Lucy and Zhang, Richard and Dekel, Tali and Isola, Phillip},
  journal={arXiv preprint arXiv:2306.09344},
  year={2023}
}

@article{wu2023hpsv2,
  title={Human preference score v2: A solid benchmark for evaluating human preferences of text-to-image synthesis},
  author={Wu, Xiaoshi and Hao, Yiming and Sun, Keqiang and Chen, Yixiong and Zhu, Feng and Zhao, Rui and Li, Hongsheng},
  journal={arXiv preprint arXiv:2306.09341},
  year={2023}
}

@inproceedings{hessel2021clipscore,
  title={Clipscore: A reference-free evaluation metric for image captioning},
  author={Hessel, Jack and Holtzman, Ari and Forbes, Maxwell and Le Bras, Ronan and Choi, Yejin},
  booktitle={Proceedings of the 2021 conference on empirical methods in natural language processing},
  pages={7514--7528},
  year={2021}
}

@article{kirstain2023pick,
  title={Pick-a-pic: An open dataset of user preferences for text-to-image generation},
  author={Kirstain, Yuval and Polyak, Adam and Singer, Uriel and Matiana, Shahbuland and Penna, Joe and Levy, Omer},
  journal={Advances in neural information processing systems},
  volume={36},
  pages={36652--36663},
  year={2023}
}

@article{you1999blind,
  title={Blind image restoration by anisotropic regularization},
  author={You, Yu-Li and Kaveh, Mostafa},
  journal={IEEE Transactions on Image Processing},
  volume={8},
  number={3},
  pages={396--407},
  year={1999},
  publisher={IEEE}
}

@article{kundur2002blind,
  title={Blind image deconvolution},
  author={Kundur, Deepa and Hatzinakos, Dimitrios},
  journal={IEEE signal processing magazine},
  volume={13},
  number={3},
  pages={43--64},
  year={2002},
  publisher={IEEE}
}

@article{yue2024deep,
  title={Deep variational network toward blind image restoration},
  author={Yue, Zongsheng and Yong, Hongwei and Zhao, Qian and Zhang, Lei and Meng, Deyu and Wong, Kwan-Yee K},
  journal={IEEE Transactions on Pattern Analysis and Machine Intelligence},
  volume={46},
  number={11},
  pages={7011--7026},
  year={2024},
  publisher={IEEE}
}

@inproceedings{ji2020realsr,
  title={Real-world super-resolution via kernel estimation and noise injection},
  author={Ji, Xiaozhong and Cao, Yun and Tai, Ying and Wang, Chengjie and Li, Jilin and Huang, Feiyue},
  booktitle={proceedings of the IEEE/CVF conference on computer vision and pattern recognition workshops},
  pages={466--467},
  year={2020}
}

@inproceedings{wang2018esrgan,
  title={Esrgan: Enhanced super-resolution generative adversarial networks},
  author={Wang, Xintao and Yu, Ke and Wu, Shixiang and Gu, Jinjin and Liu, Yihao and Dong, Chao and Qiao, Yu and Change Loy, Chen},
  booktitle={Proceedings of the European conference on computer vision (ECCV) workshops},
  pages={0--0},
  year={2018}
}

@article{kingma2013vae,
  title={Auto-encoding variational bayes},
  author={Kingma, Diederik P and Welling, Max},
  journal={arXiv preprint arXiv:1312.6114},
  year={2013}
}

@article{song2019score-matching,
  title={Generative modeling by estimating gradients of the data distribution},
  author={Song, Yang and Ermon, Stefano},
  journal={Advances in neural information processing systems},
  volume={32},
  year={2019}
}

@article{rafailov2023dpo,
  title={Direct preference optimization: Your language model is secretly a reward model},
  author={Rafailov, Rafael and Sharma, Archit and Mitchell, Eric and Manning, Christopher D and Ermon, Stefano and Finn, Chelsea},
  journal={Advances in neural information processing systems},
  volume={36},
  pages={53728--53741},
  year={2023}
}

@article{shao2024grpo,
  title={Deepseekmath: Pushing the limits of mathematical reasoning in open language models},
  author={Shao, Zhihong and Wang, Peiyi and Zhu, Qihao and Xu, Runxin and Song, Junxiao and Bi, Xiao and Zhang, Haowei and Zhang, Mingchuan and Li, YK and Wu, Yang and others},
  journal={arXiv preprint arXiv:2402.03300},
  year={2024}
}

@article{ouyang2022rlhf,
  title={Training language models to follow instructions with human feedback},
  author={Ouyang, Long and Wu, Jeffrey and Jiang, Xu and Almeida, Diogo and Wainwright, Carroll and Mishkin, Pamela and Zhang, Chong and Agarwal, Sandhini and Slama, Katarina and Ray, Alex and others},
  journal={Advances in neural information processing systems},
  volume={35},
  pages={27730--27744},
  year={2022}
}

@article{liu2025flowgrpo,
  title={Flow-grpo: Training flow matching models via online rl},
  author={Liu, Jie and Liu, Gongye and Liang, Jiajun and Li, Yangguang and Liu, Jiaheng and Wang, Xintao and Wan, Pengfei and Zhang, Di and Ouyang, Wanli},
  journal={arXiv preprint arXiv:2505.05470},
  year={2025}
}

@article{lecun2006tutorial,
  title={A tutorial on energy-based learning},
  author={LeCun, Yann and Chopra, Sumit and Hadsell, Raia and Ranzato, M and Huang, F},
  journal={Predicting structured data},
  volume={1},
  number={0},
  year={2006},
  publisher={MIT Press}
}

@article{pan2025metaquery,
  title={Transfer between modalities with metaqueries},
  author={Pan, Xichen and Shukla, Satya Narayan and Singh, Aashu and Zhao, Zhuokai and Mishra, Shlok Kumar and Wang, Jialiang and Xu, Zhiyang and Chen, Jiuhai and Li, Kunpeng and Juefei-Xu, Felix and others},
  journal={arXiv preprint arXiv:2504.06256},
  year={2025}
}

@article{han2025tar,
  title={Vision as a Dialect: Unifying Visual Understanding and Generation via Text-Aligned Representations},
  author={Han, Jiaming and Chen, Hao and Zhao, Yang and Wang, Hanyu and Zhao, Qi and Yang, Ziyan and He, Hao and Yue, Xiangyu and Jiang, Lu},
  journal={arXiv preprint arXiv:2506.18898},
  year={2025}
}

@article{qiao2025realsr-r1,
  title={RealSR-R1: Reinforcement Learning for Real-World Image Super-Resolution with Vision-Language Chain-of-Thought},
  author={Qiao, Junbo and Cai, Miaomiao and Li, Wei and Liu, Yutong and Huang, Xudong and He, Gaoqi and Xie, Jiao and Hu, Jie and Chen, Xinghao and Lin, Shaohui},
  journal={arXiv preprint arXiv:2506.16796},
  year={2025}
}

@article{wei2025pure,
  title={Perceive, Understand and Restore: Real-World Image Super-Resolution with Autoregressive Multimodal Generative Models},
  author={Wei, Hongyang and Liu, Shuaizheng and Yuan, Chun and Zhang, Lei},
  journal={arXiv preprint arXiv:2503.11073},
  year={2025}
}

@article{hu2025mvar,
  title={Auto-Regressively Generating Multi-View Consistent Images},
  author={Hu, JiaKui and Yang, Yuxiao and Liu, Jialun and Wu, Jinbo and Zhao, Chen and Lu, Yanye},
  journal={arXiv preprint arXiv:2506.18527},
  year={2025}
}

@article{hu2025dcpt,
  title={Universal Image Restoration Pre-training via Masked Degradation Classification},
  author={Hu, JiaKui and Yao, Zhengjian and Jin, Lujia and Chen, Yinghao and Lu, Yanye},
  journal={arXiv preprint arXiv:2510.13282},
  year={2025}
}

@article{hu2025omni-view,
  title={Omni-View: Unlocking How Generation Facilitates Understanding in Unified 3D Model based on Multiview images},
  author={Hu, JiaKui and Zhao, Shanshan and Chen, Qing-Guo and Qiu, Xuerui and Liu, Jialun and Xu, Zhao and Luo, Weihua and Zhang, Kaifu and Lu, Yanye},
  journal={arXiv preprint arXiv:2511.07222},
  year={2025}
}

@article{lin2025uniworld,
  title={Uniworld: High-resolution semantic encoders for unified visual understanding and generation},
  author={Lin, Bin and Li, Zongjian and Cheng, Xinhua and Niu, Yuwei and Ye, Yang and He, Xianyi and Yuan, Shenghai and Yu, Wangbo and Wang, Shaodong and Ge, Yunyang and others},
  journal={arXiv preprint arXiv:2506.03147},
  year={2025}
}

@article{jiang2025survey,
  title={A survey on all-in-one image restoration: Taxonomy, evaluation and future trends},
  author={Jiang, Junjun and Zuo, Zengyuan and Wu, Gang and Jiang, Kui and Liu, Xianming},
  journal={IEEE Transactions on Pattern Analysis and Machine Intelligence},
  year={2025},
  publisher={IEEE}
}

@article{chen2025nft,
  title={Bridging supervised learning and reinforcement learning in math reasoning},
  author={Chen, Huayu and Zheng, Kaiwen and Zhang, Qinsheng and Cui, Ganqu and Cui, Yin and Ye, Haotian and Lin, Tsung-Yi and Liu, Ming-Yu and Zhu, Jun and Wang, Haoxiang},
  journal={arXiv preprint arXiv:2505.18116},
  year={2025}
}

@article{yang2025sodiff,
  title={{SODiff}: Semantic-Oriented Diffusion Model for {JPEG} Compression Artifacts Removal},
  author={Yang, Tingyu and Gong, Jue and Guo, Jinpei and Li, Wenbo and Guo, Yong and Zhang, Yulun},
  journal={arXiv preprint arXiv:2508.07346},
  year={2025}
}

@inproceedings{qi2024spire,
  title={{SPIRE}: Semantic Prompt-Driven Image Restoration},
  author={Qi, Chenyang and Tu, Zhengzhong and Ye, Keren and Delbracio, Mauricio and Milanfar, Peyman and Chen, Qifeng and Talebi, Hossein},
  booktitle={Proceedings of the European Conference on Computer Vision (ECCV)},
  year={2024}
}

@inproceedings{yu2024promptfix,
  title={{PromptFix}: You Prompt and We Fix the Photo},
  author={Yu, Yongsheng and Zeng, Ziyun and Hua, Hang and Fu, Jianlong and Luo, Jiebo},
  booktitle={Advances in Neural Information Processing Systems (NeurIPS)},
  year={2024}
}

\vspace{-15 mm}
\begin{IEEEbiography}[{\includegraphics[width=1in,height=1.25in, clip,keepaspectratio]{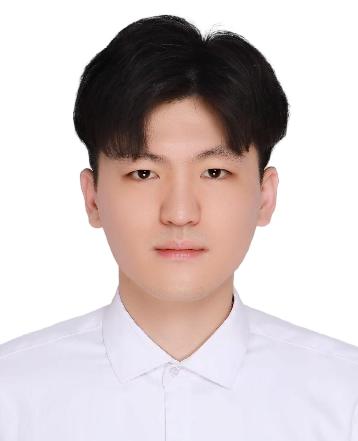}}]{Zhengjian Yao} received the B.S. degree in the School of Mathematics and Statistics from Xi'an Jiaotong University in 2022. He is currently pursuing the Ph.D. degree at the Medical Intelligence Lab, Peking University. His current research interests include low-level vision, applications of image generation, and reinforcement learning theories for large language models.
\end{IEEEbiography}\vspace{-15 mm}

\begin{IEEEbiography}[{\includegraphics[width=1in,height=1.25in, clip,keepaspectratio]{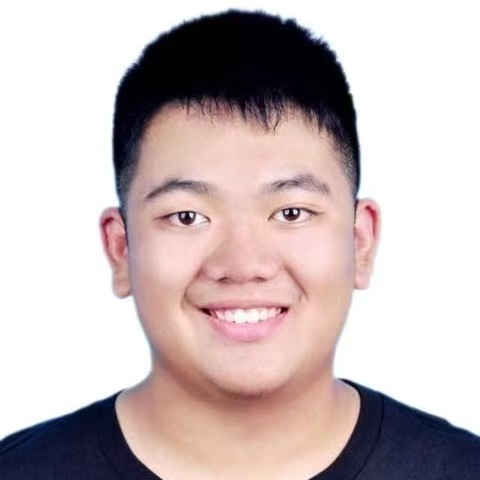}}]
{JiaKui Hu} received the B.S. degree in School of Physics and Optoelectronic Engineering from Xidian University in 2023. He is currently pursuing the Ph.D. degree at Medical Intelligence Lab, Peking University. His current research interests include low-level vision and unified model.
\end{IEEEbiography}\vspace{-15 mm}

\begin{IEEEbiography}[{\includegraphics[width=1in,height=1.25in, clip,keepaspectratio]{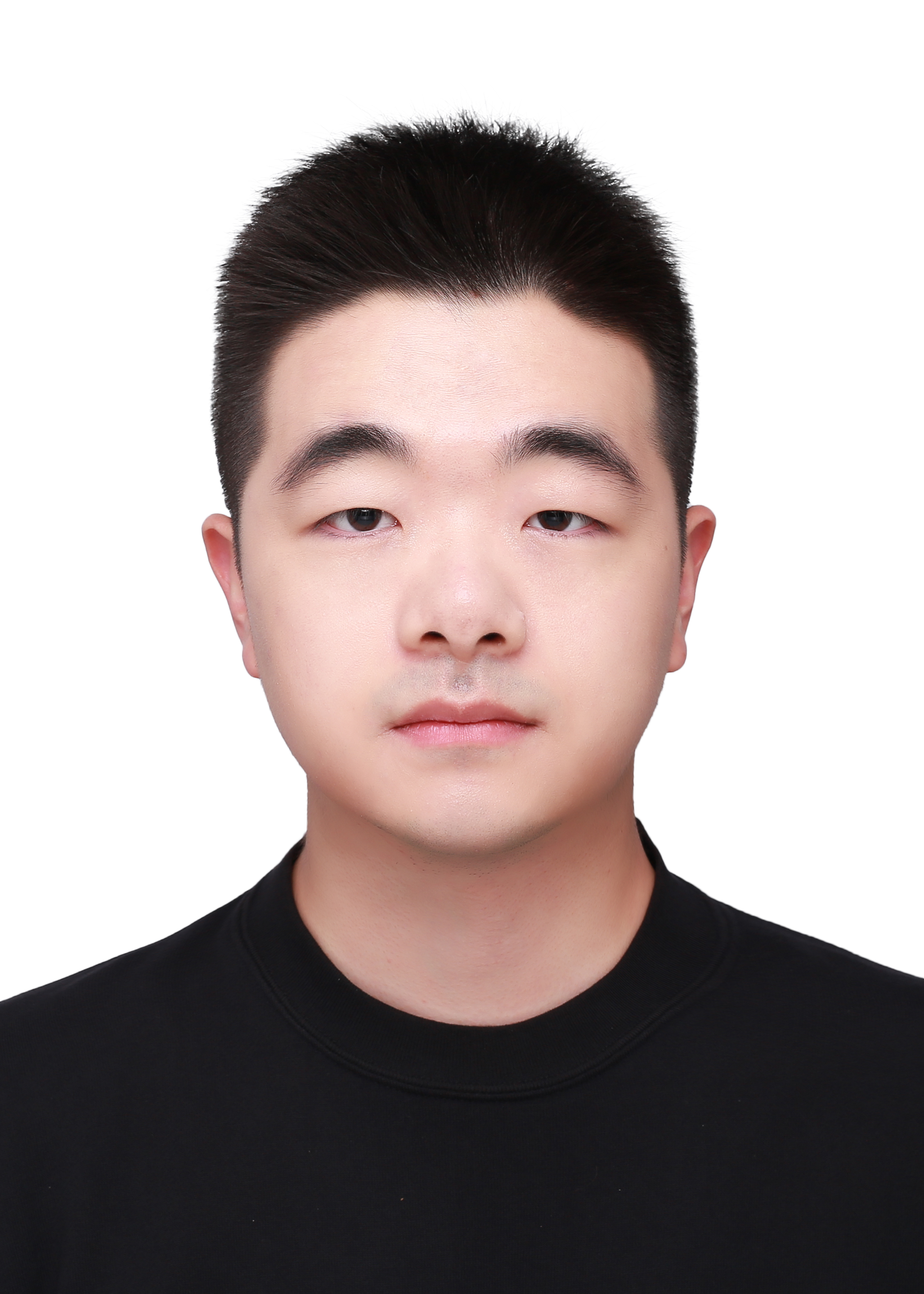}}]
{Kaiwen Li} received the Bachelor’s degree in Electronic and Information Engineering from China University of Petroleum, Qingdao, China in 2021 and the Master’s degree in Electronic science and technology from University of Electronic Science and Technology of China, Chengdu, China in 2024. He is currently pursuing the Ph.D. degree at Medical Intelligence Lab, Peking University. His current research interests include weakly supervised learning, image generation, and multimodal large language models.
\end{IEEEbiography}\vspace{-15 mm}

\clearpage
\begin{IEEEbiography}[{\includegraphics[width=1in,height=1.25in, clip,keepaspectratio]{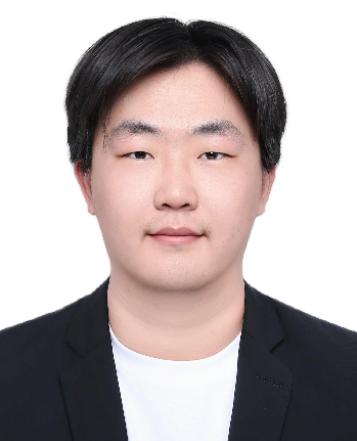}}]{Hangzhou He} received the B.S. degree in Theoretical and Applied Mechanics from Peking University, Beijing, China in 2024. He is now a Ph.D. student majoring in Biomedical Engineering at Peking University. His research interests focus on the trustworthiness of deep learning models, including explainability, generalization and their applications in medical image analysis. 
\end{IEEEbiography}\vspace{-15 mm}

\begin{IEEEbiography}[{\includegraphics[width=1in,height=1.25in, clip,keepaspectratio]{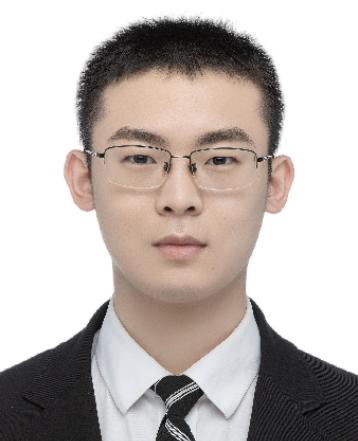}}]{Xinliang Zhang} received the B.S. degree in Electronic Information of Engineering from Ocean University of China, Qingdao, China in 2021 and the Master degree in Computer Science and Technology from Tianjin University in 2024, Tianjin, China. He is now pursuing the Ph.D. degree at Medical Intelligence Lab, Peking University. His research interests include computer vision, deeplearning, and medical image analysis.
\end{IEEEbiography}\vspace{-15 mm}

\begin{IEEEbiography}[{\includegraphics[width=1in,height=1.25in, clip,keepaspectratio]{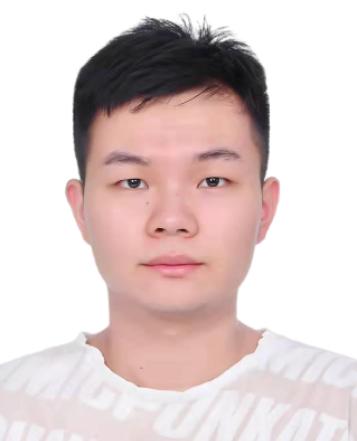}}]{Shuang Zeng} received a bachelor's degree in Engineering from Peking University, Beijing, China in 2021. He is currently a joint Ph.D. student of Peking University - Georgia Institute of Technology - Emory University Biomedical Engineering Program. His research mainly focuses on self-supervised contrastive learning and medical image processing.
\end{IEEEbiography}\vspace{-15 mm}

\begin{IEEEbiography}[{\includegraphics[width=1in,height=1.25in, clip,keepaspectratio]{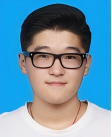}}]{Lei Zhu} received the Ph.D. degree from Peking University, Beijing, China. He is currently a postdoc researcher at Medical Intelligence Lab, Peking University. His current research interests include weakly supervised learning, multimodal large language models, and medical image processing. 
\end{IEEEbiography}\vspace{-15 mm}

\begin{IEEEbiography}[{\includegraphics[width=1in,height=1.25in, clip,keepaspectratio]{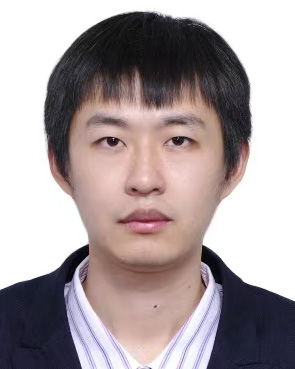}}]{Yanye Lu} is currently an Assistant Professor at Peking University. He was nominated by the Ministry of Education of China for the Young Changjiang Scholar Program in 2024. His research focuses on artificial intelligence (AI), computer vision, and multimodal medical imaging, with core interests in limited-supervision learning, robust and interpretable modeling, and multimodal generative AI. His work primarily targets the processing, analysis, and visualization of multimodal cross-scale biomedical information and medical images. He has published over 90 papers in top-tier journals across related fields (TPAMI, IJCV, TIP, TNNLS, TMI, MedIA, JNM) as well as in flagship computer science conferences including CVPR, ICCV, ICLR, NeurIPS, AAAI, and ECCV.
\end{IEEEbiography}\vspace{-15 mm}


\end{document}